\documentclass[twoside]{article}

\usepackage[accepted]{aistats2023}
\usepackage[round]{natbib}

\usepackage{mathtools}
\DeclarePairedDelimiterX{\infdivx}[2]{(}{)}{%
  #1\;\delimsize\|\;#2%
}
\newcommand{\infdiv}{\text{KL}\infdivx}
\usepackage{bm}
\usepackage{amssymb}
\DeclareMathOperator*{\argmax}{arg\,max}
\usepackage{dsfont}

\usepackage[usenames,dvipsnames]{xcolor}

\usepackage[normalem]{ulem}
\usepackage{subcaption}

\usepackage{amsthm}
\newtheorem{definition}{Definition}
\newtheorem{remark}{Remark}

\usepackage{booktabs}
\usepackage{multirow}
\usepackage{algorithm}
\usepackage{algpseudocode}
\usepackage[autostyle]{csquotes}
\MakeOuterQuote{"}

\usepackage{hyperref} 

\hypersetup{
    colorlinks,
    linkcolor={red!50!black},
    citecolor={blue!50!black},
    urlcolor={blue!80!black}
}
\begin{document}

%

%

\twocolumn[

\aistatstitle{Variational Inference for Neyman-Scott Processes}

\aistatsauthor{ Chengkuan Hong \And Christian R. Shelton }

\aistatsaddress{ Tsinghua University \\ chong009@ucr.edu \And  University of California, Riverside \\ cshelton@cs.ucr.edu } ]

\begin{abstract}
Neyman-Scott processes (NSPs) have been applied across a range of fields to
model points or temporal events with a hierarchy of clusters.  Markov chain
Monte Carlo (MCMC) is typically used for posterior sampling in the model.
However, MCMC's mixing time can cause the resulting inference to be slow, and thereby slow down model learning and prediction. We develop 
the first variational inference (VI) algorithm for NSPs,
and give two examples of suitable variational posterior point process
distributions.  Our method minimizes the inclusive Kullback-Leibler (KL)
divergence for VI to obtain the variational parameters.  We generate
samples from the approximate posterior point processes much faster than
MCMC, as we can directly estimate the approximate posterior point processes
without any MCMC steps or gradient descent.  We include synthetic and
real-world data experiments that demonstrate our VI algorithm achieves
better prediction performance than MCMC when computational time
is limited.
\end{abstract}

\section{INTRODUCTION}

Neyman-Scott processes \citep{neyman1958statistical} have achieved great success
in many fields, \textit{e.g.}, pandemic modeling \citep{park2022an}, neuroscience
\citep{williams2020point}, and seismology \citep{hong2022deep}, where posterior
sampling plays an important role. Many MCMC algorithms \citep[\textit{e.g.}, ][]{moller2003statistical,williams2020point,hong2022deep} have been proposed for the posterior sampling. However, a large number of mixing steps is generally required for MCMC, which makes it inefficient to generate samples from the posterior point process. 

To accelerate the approximate posterior inference of the hidden points, we
propose a variational inference (VI) algorithm. We employ an autoencoder
to construct two variational posterior point processes. One of these has the same functional form as the virtual point processes, which work as auxiliary variables for MCMC, in \citet{hong2022deep}, and the other uses self-attention \citep{vaswani2017attention} to construct a more general approximation. We can sample from our approximation much faster than from MCMC because mixing is not needed for the approximation. With faster sampling from our approximate posterior point processes, we are able to achieve better prediction results
when there is a time constraint.

We briefly review VI and Neyman-Scott processes in 
Sec.~\ref{sec:related-work}. Sec.~\ref{sec:apprx-posterior-pp} gives a general
formula for our approximate posterior point processes, and 
Sec.~\ref{sec:examples-for-q} gives two specific examples for the approximation.
Sec.~\ref{sec:inference} gives details about the inference procedure. 
Sec.~\ref{sec:prediction} explains how to do prediction and discusses why our
approximate posterior point processes can behave better than MCMC when there is a
constraint for the time. Finally, we use experiments to support our claim in
Sec.~\ref{sec:experiments} and conclude this paper in 
Sec.~\ref{sec:conclusion}. While our experiments are only for temporal point
processes,
our algorithms can be applied to a general spatio-temporal point process as explained in Remark \ref{remark:general-spatio-temporal}. 
\section{BACKGROUND} \label{sec:related-work}
\subsection{Variational Inference} VI transforms a posterior inference problem
into an optimization problem, and it has already been very successful for
probabilistic modeling. Usually, VI first constructs a family of
approximate distributions \(q\), and then adjusts the parameters of \(q\) to
approach the true posterior distribution \(p\) by minimizing a divergence metric.
Most VI algorithms try to minimize the exclusive KL divergence,
\(\infdiv{q}{p}\)
\cite[\textit{e.g.},][]{jordan1999introduction,kingma2014auto,ranganath2014black,blei2017variational}.
While it is computationally efficient to optimize the exclusive KL divergence, it
can lead to underestimation of the uncertainty of the posterior
\citep{naesseth2020markovian}. To mitigate the underestimation issue,
\citet{naesseth2020markovian} proposed a VI algorithm, called Markovian score
climbing (MSC), that minimizes the inclusive KL divergence, \(\infdiv{p}{q}\).
MSC uses MCMC to get an unbiased estimate of the gradient of the inclusive KL
divergence. \citet{naesseth2020markovian} also shows that MSC can be combined with
maximum likelihood estimation (MLE) to jointly learn variational parameters and
model parameters.

In a similar fashion, we design a VI algorithm for NSPs that
minimizes inclusive KL divergence.
Different from MSC, whose variational parameters are for Markov kernels, our
variational parameters are for auxiliary variables of MCMC. Moreover, the number
of dimensions is fixed in \citet{naesseth2020markovian}, and our MCMC has an
unbounded number of dimensions. To the best of our knowledge, we are the first to design a VI algorithm for NSPs and there does not exist any pre-existing work for minimizing the exclusive KL divergence. While we only focus on the VI for minimizing inclusive KL divergence in this paper, the minimization of exclusive KL divergence could also be interesting and we leave it for future research. 

\subsection{Neyman-Scott Processes} The Neyman-Scott process (NSP) was first
proposed by \citet{neyman1958statistical}. NSPs are a class of hierarchical point
process models built by stacking Poisson processes into a network. NSPs were
originally univariate, \textit{i.e.}, can only be applied to events with one type
or mark. \citet{hong2022deep} introduce multivariate NSPs with hierarchical structures, called deep
Neyman-Scott processes (DNSPs). We use "deep" to differentiate our work from other work that only has one hidden layer \citep[\textit{e.g.},][]{moller2003statistical,linderman2017bayesian,williams2020point}.  While \citet{hong2022deep} only have experiments for temporal point processes (TPPs), their algorithms can be
used for spatio-temporal NSPs with multiple dimensions. 

The major assumption of DNSPs is that the observed events (points) are
triggered by the hidden events (points). For example, earthquakes are
usually triggered by sudden energy releases of the Earth. The times and
locations of energy releases are hidden (not directly observed) points.
We can leverage the power of DNSPs to infer the potential times and
locations of energy releases given the times, locations, and
magnitudes of the observed earthquakes. Then, based on the inferred
hidden information, we can predict the time and the magnitude of the
next future earthquake conditional on the observed earthquakes.   

More generally, the hierarchical systems of DNSPs are formed by the
iterative generation procedures of the hidden points. Each hidden point
(parent) can trigger a set of hidden points as its children. The
generated child points can also work as parent points and trigger their
own sets of children. This iterative process continues until the
observed points are generated. With this hierarchical structure, we are
able to infer more hidden information. For example, we can infer the
hidden triggers of the energy releases of earthquakes. The experiments
in \citet{hong2022deep} and our experiments in
Sec.~\ref{sec:real-world-experiments} also show that more hidden point
processes can bring better prediction performance.

We typically use conditional intensity functions (CIFs) to describe TPPs.
\begin{definition}[conditional intensity function] \label{def:cif} The {\em conditional
	intensity function} (CIF) of a TPP is defined as 
\begin{align*}
	\lambda(t)=\lim_{\Delta t\rightarrow 0}\frac{\text{\em Pr}(\text{\em One event in }[t,t+\Delta t]\ \mid\ \mathcal{H}_t)}{\Delta t},
\end{align*}
where \(\mathcal{H}_t\) represents the events that happened before \(t\).
\end{definition}
The log-likelihood function for a TPP defined on \((0,T]\) is 
\begin{equation*}
    \sum_{i=1}^n\log \lambda(t_i)-\int_0^T\lambda(t)dt,
\end{equation*}
where \(\lambda(t)\) is the CIF and \(t_1 \leq t_2 \leq t_3 \leq \cdots \leq
t_n\) are the times of observations.

Figs.~1 and 2 in \citet{hong2022deep} provide good illustrations for deep Neyman-Scott processes, where each observed sequence of events has \(L\) hidden
layers of (multiple) Poisson processes. The CIF for each Poisson process of layer
$l$ is determined by the events from the layer immediately above layer $l+1$. For simplicity, we omit the data index \(n\) in the following formal description of the generative process for DNSPs, and
only consider TPPs in this paper. Let \(Z_{\ell,k}\) represent the \(k\)-th hidden Poisson process at layer \(\ell\), \(K_\ell\) be the number of point processes at layer \(\ell\), \(\phi_{\bm{\theta}_{(\ell+1,i) \rightarrow (\ell,k)}}\) be the kernel function that connects \(Z_{\ell,k}\) and \(Z_{\ell+1,i}\), \(\mathbf{z}_{\ell,i} =
\{t_{\ell,i,j}\}_{j=1}^{m_{\ell,i}}\)
be a realization for  \(Z_{\ell,i}\), \(\mathbf{z}_{\ell}=\{\mathbf{z}_{\ell,i}\}_{i=1}^{K_\ell}\), and \(\mathbf{z}_{0}=\mathbf{x}\). To generate the events for the bottom layer, we first need to generate random events from the top layer. The CIF for \(Z_{L,k}\) is
\begin{equation*}
    \lambda_{L,k}(t) = \mu_k\text{, where }\mu_k > 0.
\end{equation*}
Then, we can draw samples for each hidden Poisson process conditional on the Poisson processes that are from the layers immediately above. The CIF 
for \(Z_{\ell,k}\) is
\begin{equation}
    \lambda_{\ell,k}(t)=\sum_{i=1}^{K_{\ell+1}}\sum_{t_{\ell+1,i,j}}\phi_{\bm{\theta}_{(\ell+1,i) \rightarrow (\ell,k)}}(t-t_{\ell+1,i,j}). \label{eq:real-CIF}
\end{equation}
We denote this point process as the real point process (RPP), to make a distinction 
with the virtual point process (VPP), below, used to add auxiliary variables to
a Markov chain.

\subsection{Kernel Function} \label{sec:kernel-function}
We choose the kernel function to be a Weibull kernel,
\begin{equation*}
    \phi_{\bm{\theta}}(x) = \begin{cases}
    p \cdot
	    \frac{k}{\lambda}\left(\frac{x}{\lambda}\right)^{k-1}e^{-(x/\lambda)^k}
	    & \text{for }x>0,\ k,\lambda > 0,\\
	    0 &\text{for }x \leq 0,
    \end{cases}
\end{equation*}
where \(\bm{\theta}=\{p,k,\lambda\}\). Here, we omit the subscripts of \(\bm{\theta}\) in the definition for simplicity as it is the general form. Similar to the gamma kernel
used in previous work,
the Weibull kernel converges to 0 when \(x\) goes to infinity as the influence of
the events from the past will fade eventually. The shape of a Weibull kernel is
very similar to a gamma kernel, and, with different combinations of the
parameters, the Weibull kernel can also monotonically decrease or behave like a
Gaussian function. But, the Weibull kernel has the advantage over the gamma kernel that the gradients of the Weibull kernel function itself and the integral of the Weibull kernel function are both analytically available. The only restriction for the kernel function is that Eqs.~\ref{eq:real-CIF}, \ref{eq:virtual-CIF}, \ref{eq:bnsp-cif}, and \ref{eq:bsap-cif} are non-negative.

\subsection{Monte Carlo Expectation-Maximization} \citet{hong2022deep} introduced a Monte Carlo expectation-maximization (MCEM)
algorithm for inference.  

Virtual events work as auxiliary variables to
accelerate the mixing of MCMC chains, and are candidate events for real events. When searching for the positions of the real events, we only need to search where the virtual events appear instead of in the whole space. Because of this, the virtual point processes (VPPs)
\(\tilde{\mathbf{Z}}\) are designed to be as close as possible to the true
posterior point processes of the real point processes \(\mathbf{Z}\). For the same reason, the CIFs
for VPPs evolve in the reverse directions (temporally backward and ``up''
the layers) relative to the real point processes (temporally forward and ``down''
the layers). 

Similar to the RPPs, \(\tilde{Z}_{\ell,k}\) represents the \(k\)-th
hidden Poisson process on layer \(\ell\),
\(\tilde{\phi}_{\tilde{\bm{\theta}}_{(\ell-1,i) \rightarrow (\ell,k)}}\) is the
kernel function connects \(Z_{\ell-1,i}\) and \(\tilde{Z}_{\ell,k}\), and
\(\tilde{\mu}_{\ell,k} \geq 0\) is the base rate. In this case, the history \(\mathcal{H}_t\) in Definition \ref{def:cif} becomes the events that happened after \(t\). Then, the CIF for
\(\tilde{Z}_{\ell,k}\) conditioned on \(\mathbf{Z}_{\ell-1}\) is 
\begin{equation}
	\tilde{\lambda}_{\ell,k}(\tilde{t}) = 
	\tilde{\mu}_{\ell,k}+\!\sum_{i=1}^{K_{\ell-1}}\!\sum_{t_{\ell-1,i,j}}\!\!\tilde{\phi}_{\tilde{\bm{\theta}}_{(\ell-1,i) \rightarrow (\ell,k)}}(t_{\ell-1,i,j}-\tilde{t}). \label{eq:virtual-CIF}
\end{equation}
We briefly outline the MCEM algorithm \citep{hong2022deep} in Alg.~\ref{alg:mcem}. 
\(\log p(\mathbf{x}, \mathbf{z})\) in line
\ref{alg:mcem-opt-marginal} is the joint log-likelihood of the hidden RPPs and
the observation \(\mathbf{x}\). \(\log q(\mathbf{z})\) in line
\ref{alg:mcem-opt-kl} represents the log-likelihood for the VPPs.  There is a
slight difference between Alg.~\ref{alg:mcem} and the employed MCEM algorithm: We
directly maximize the constant rates \(\mu\) instead of doing gradient
ascent. Line \ref{alg:mcem-mcmc} does posterior sampling via MCMC, and the
previous sample \(\mathbf{z}_{n-1}\) serves as the initial state of the next MCMC
step. Line \ref{alg:mcem-opt-marginal} is used to maximize the marginal
likelihood. Line \ref{alg:mcem-opt-kl} adjusts the parameters for VPPs to make
VPPs closer to the posterior RPPs. \(\eta_n\) and \(\tilde{\eta}_n\) are step sizes for gradient ascent.

\citet{hong2022deep} only use Alg.~\ref{alg:mcem} to learn the model parameters for DNSPs, without any discussion of the variational inference. However, we can directly use Alg.~\ref{alg:mcem} to learn the parameters for our approximate point processes, which is one of our main contributions. 
\begin{algorithm}[tb]
\caption{MCEM for DNSPs}
\label{alg:mcem}
\textbf{Input}: data \(\mathbf{x}\), model \(\mathcal{M}\)\\
\textbf{Initialization}: parameters for RPPs \(\bm{\Theta}_0\), parameters for VPPs \(\tilde{\bm{\Theta}}_0\), initial sample for RPPs \(\mathbf{z}_0\), and iterations \(N\).\\
\textbf{Output}: \(\bm{\Theta}_N \approx \bm{\Theta}^*\), \(\tilde{\bm{\Theta}}_N \approx \tilde{\bm{\Theta}}^*\)
\begin{algorithmic}[1] 
\For{\(n = 1\) \textbf{to} \(N\)}
\State sample \(\mathbf{z}_n \sim  p\left(\mathbf{z} \mid \mathbf{x}; \bm{\Theta}_{n-1},\mathbf{z}_{n-1}\right)\) \label{alg:mcem-mcmc}
\State \(\bm{\Theta}_n \leftarrow \bm{\Theta}_{n-1} + \eta_n\nabla_{\bm{\Theta}}\log p(\mathbf{x}, \mathbf{z}; \bm{\Theta}_{n-1})\) \label{alg:mcem-opt-marginal}
\State \(\tilde{\bm{\Theta}}_n \leftarrow \tilde{\bm{\Theta}}_{n-1} + \tilde{\eta}_n\nabla_{\tilde{\bm{\Theta}}}\log q(\mathbf{z};\mathbf{x},  \tilde{\bm{\Theta}}_{n-1})\) \label{alg:mcem-opt-kl}
\EndFor
\end{algorithmic}
\end{algorithm}
\begin{figure}[tb]
\centering
\begin{subfigure}{.4\textwidth}
    \centering
    \includegraphics[width=\textwidth]{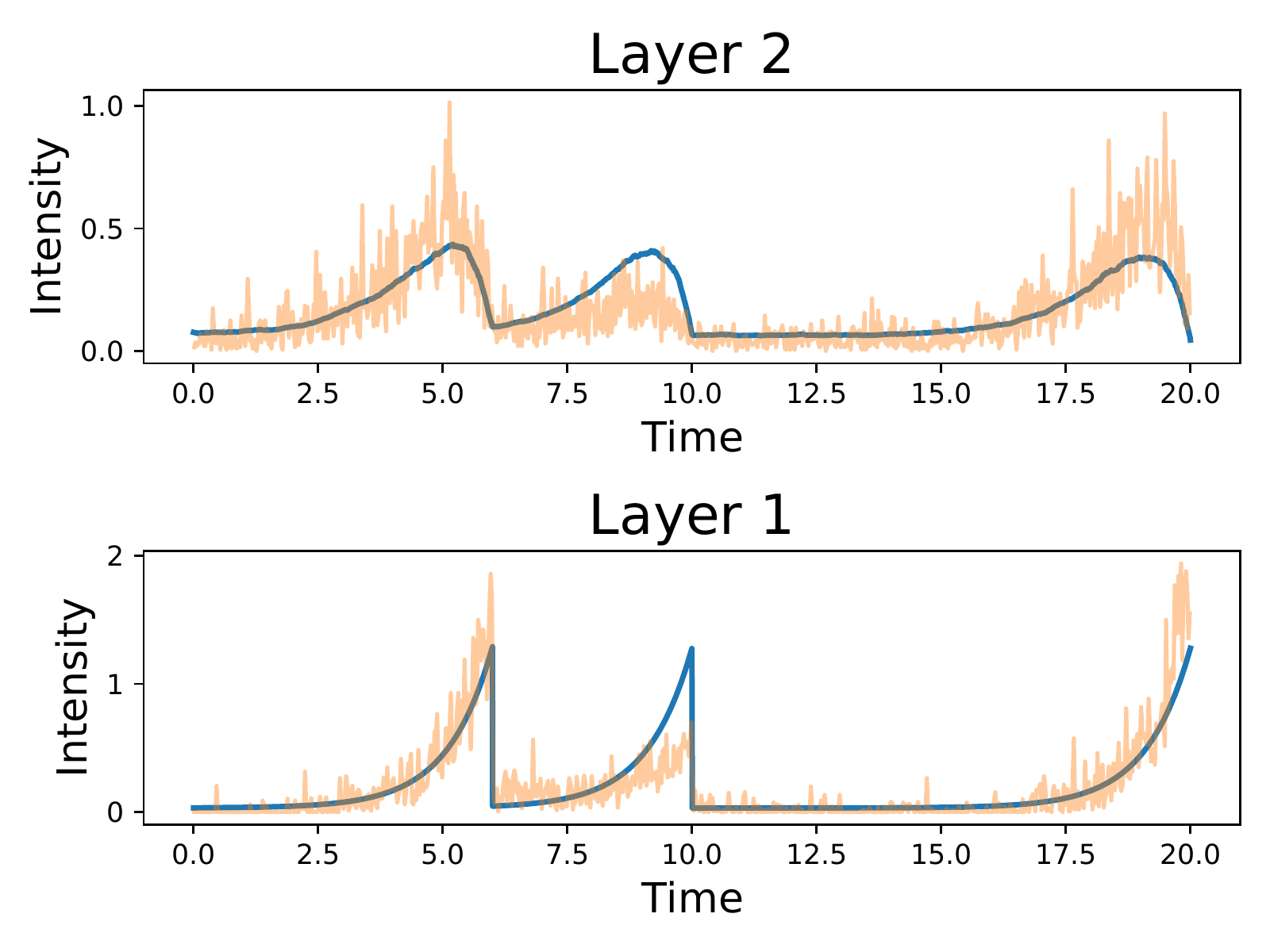}
    \caption{UNSP vs. MCMC}
    \label{fig:bnsp}         
\end{subfigure}
\vskip .2in
\begin{subfigure}{.4\textwidth}
         \centering
    \includegraphics[width=\textwidth]{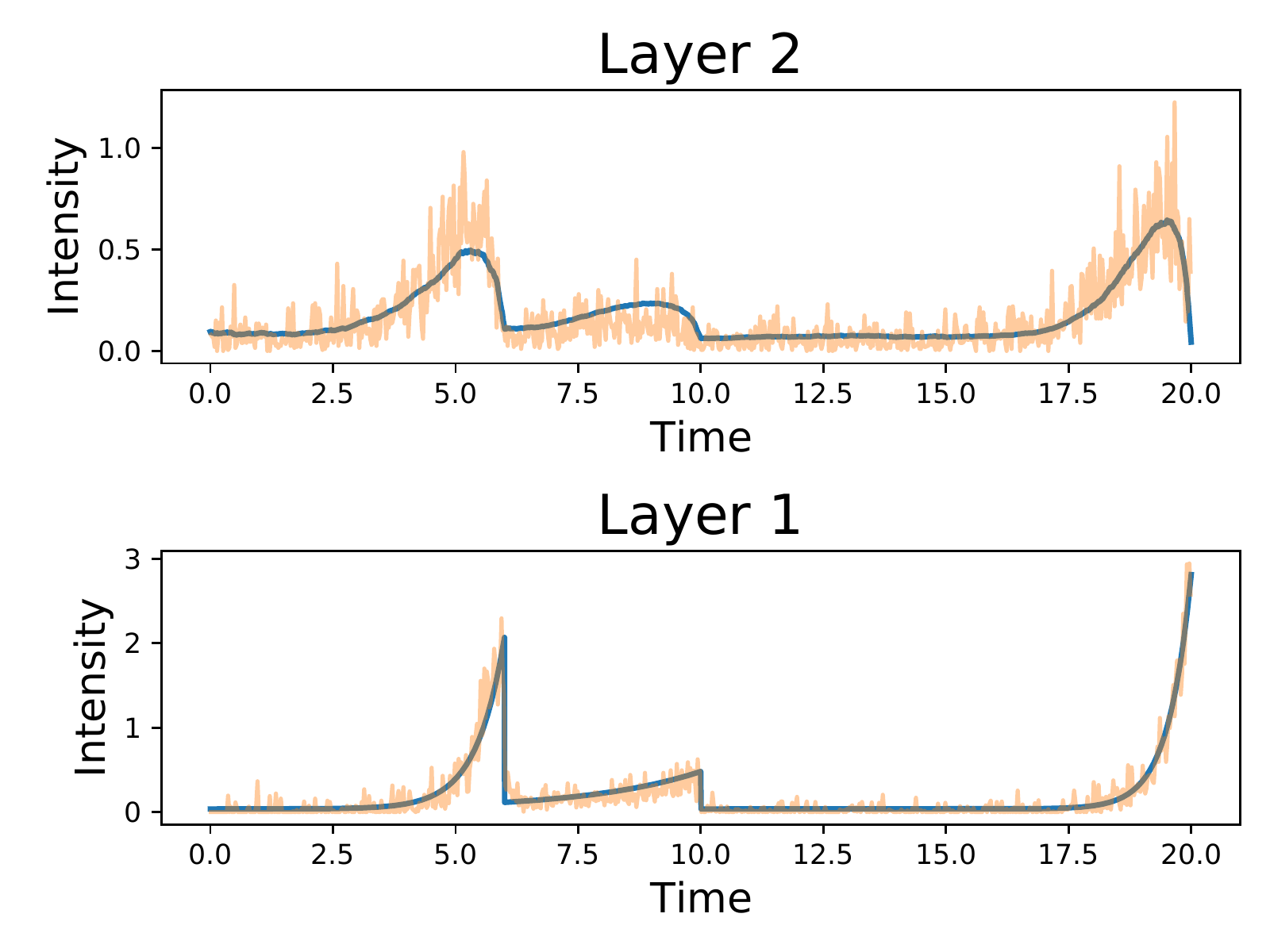}
    \caption{USAP vs. MCMC}
    \label{fig:bsap}
\end{subfigure}
\caption{Comparison Between UNSP and USAP}
\label{fig:mcmc-vs-vi}
\end{figure}
\section{APPROXIMATE POSTERIOR POINT PROCESSES} \label{sec:apprx-posterior-pp}
We denote our variational approximate point processes as
\(\mathbf{Z}^{I}\), and we assume they have hierarchical structures like DNSPs.
However, the approximate point processes are generated from bottom to top
(upward), instead of from top to bottom (downward). The upward approximation
mechanism allows us to infer the approximate posterior distributions of the point processes directly from the observation, which can help accelerate the sampling process for the posterior point processes. Moreover, the events from the layers immediately below can give heuristics for the positions of the posterior events from the layers immediately above.
\paragraph{Generative Semantics for \(\mathbf{Z}^{I}\)}
The approximate posterior point processes are generative models, and they are denoted as \(\mathbf{Z}^{I}=\{\mathbf{Z}^{I}_1,\cdots,\mathbf{Z}^{I}_L\}\), with \(\mathbf{Z}^{I}_\ell=\{Z^{I}_{\ell,k}\}_{k=1}^{K_\ell}\).

\(\mathbf{Z}^{I}_1\) is assumed to be Poisson processes, and the CIF for \(Z^{I}_{1,k}\) is \begin{equation}
    \lambda^{I}_{1,k}(t) = q^I_{1,k}(t;\mathbf{z}^I_{0}=\mathbf{x},\bm{\theta}^{I}_{1,k}(t)),
\end{equation}
where \(q^I_{1,k}(t;\mathbf{x}, \bm{\theta}^{I}_{1,k}(t))\) is a function of
\(t\) with some parameters determined by the events from the observation
\(\mathbf{x}\), and other parameters \(\bm{\theta}^{I}_{1,k}(t)\) that can depend
on time \(t\).

Next, we draw samples for each \(\mathbf{Z}^{I}_\ell\) conditional on the samples \(\mathbf{z}^{I}_{\ell-1}\) from \(\mathbf{Z}^{I}_{\ell-1}\). The CIF for \(Z^{I}_{\ell,k}\) conditional on \(\mathbf{Z}^{I}_{\ell-1}\) is 
\begin{equation}
     \lambda^{I}_{\ell,k}(t) = q^I_{\ell,k}(t;\mathbf{z}^{I}_{\ell-1},\bm{\theta}^{I}_{\ell,k}(t)),
\end{equation}
where \(q^I_{\ell,k}(t;\mathbf{z}^{I}_{\ell-1},\bm{\theta}^{I}_{\ell,k}(t))\) is
a function of \(t\) with some parameters determined by the events from
\(\mathbf{z}^{I}_{\ell-1}\), and other parameters can also depend on time \(t\).

\section{EXAMPLES FOR VARIATIONAL POINT PROCESSES} \label{sec:examples-for-q}
As explained above, there are two major goals when constructing the
approximations: (1) the approximations should be able to propagate the
information from the observations to the top, and (2) the approximate posterior
point processes should be close to the true posterior point processes. For these
purposes, we consider the following two examples of functional forms for
\(q^I_{\ell,k}(\cdot)\) in this paper: upward Neyman-Scott processes and upward self-attention processes. 

\paragraph{Upward Neyman-Scott Processes (UNSPs)}

Like the virtual point processes, we can assume the approximate point processes
are NSPs evolving in an upward direction. In this case, the CIF is 
\begin{align}
    \!\!\!\!\lambda^{I}_{\ell,k}(t)\! &=\! q^I_{\ell,k}\left(t;\bm{z}^I_{\ell-1},\bm{\theta}^{I}_{\ell,k}  \right)\nonumber \\
    &=\!{\mu}^{I}_{\ell,k}+\!\sum_{i=1}^{K_{\ell-1}}\!\sum_{t_{\ell-1,i,j}}\!\!\phi_{\bm{\theta}^{I}_{(\ell-1,i) \rightarrow (\ell,k)}}(t_{\ell-1,i,j}-t), \label{eq:bnsp-cif}
\end{align}
where
\(\bm{\Theta}^{I}_{\ell,k}=\left\{\mu_{\ell,k}^I,\left\{\bm{\theta}^{I}_{(\ell-1,i)\rightarrow
(\ell,k)}\right\}_{i=1}^{K_{\ell-1}}\right\}\) and \({\mu}^{I}_{\ell,k} \geq 0\).
We see that Eq.~\ref{eq:bnsp-cif} has the same functional form as Eq.~\ref{eq:virtual-CIF}.

\paragraph{Upward Self-Attention Processes (USAPs)}
Self-attention has been widely used in the modeling for point processes \citep[\textit{e.g.},][]{zuo2020transformer,zhang2020self,chen2021neural}. Here, we use self-attention to encode the event information from a layer below to a layer above. Each event \(t_{\ell-1,i,j}\) is encoded into a hidden vector \(\bm{h}_{\ell-1,i,j}=f_{\text{Attn}}(t_{\ell-1,i,j}, \mathbf{z}^{I}_{\ell-1}=\{\{t_{\ell-1,i,j}\}_{j=1}^{m_{\ell-1,i}}\}_{i=1}^{K_{\ell-1}}\})\) through self-attention.
The CIF for \(t_{\ell-1,i,j-1} < t \leq t_{\ell-1,i,j}\) becomes
\begin{align}
    \lambda^{I}_{\ell,k}(t) &= q^I_{\ell,k}\left(t;\mathbf{z}^I_{\ell-1}, \bm{\theta}^{I}_{\ell,k}\right)\nonumber\\
    &= {\mu}^{I}_{\ell,k} + \phi_{\bm{\theta}^{I}_{\ell-1,i,j}}(t_{\ell-1,i,j}-t), \label{eq:bsap-cif}
\end{align}
where \(\bm{\Theta}^{I}_{\ell,k}=\left\{\mu^I_{\ell,k},  \left\{\left\{\bm{\theta}^I_{\ell-1,i,j}\right\}_{i=1}^{K_{\ell-1}}\right\}_{j=1}^{m_{\ell-1,i}}\right\}\), \({\mu}^{I}_{\ell,k} \geq 0\), and \(\bm{\theta}^{I}_{\ell,k}\) is the output of a linear transformation of the hidden vector \(\bm{h}_{\ell-1,i,j}\). More details can be found in Appx.~\ref{sec:self-attention-details}.


\paragraph{Comparison}
We plot the intensity functions estimated by using UNSPs, USAPs, and MCMC in Fig.~\ref{fig:mcmc-vs-vi} to compare the approximation abilities of UNSPs and USAPs. We choose 3 events by hand in the observation (times at 6, 10, and 20) to illustrate the VI effects in a clean and typical example, and we use MCMC, UNSPs, and USAPs to infer the posterior point processes. 

The model parameters for the underlying DNSP are fixed. We only adjust the variational parameters to let the variational point processes (USAPs and UNSPs) approximate the posterior point processes. Layer 2 is further away from the observation than layer 1. The yellow lines
with many spikes are the approximate intensity functions estimated by using the
samples from MCMC. We divide the time interval into many small subintervals, and
for each small subinterval, the approximate intensity function is the number of
events in that small subinterval divided by the length of that small subinterval.
The blue solid lines are approximate intensity functions for VI. We obtain the
intensity function for layer 1 directly from \(q^I_{1,0}(\cdot)\). For the
approximate intensity function for layer 2, we first generate many 
samples from \(q^I_{1,0}(\cdot)\) which induce samples of
\(q^I_{2,0}(\cdot)\). 
Then the approximate intensity function for layer 2 is the mean of all the samples for
\(q^I_{2,0}(\cdot)\). We see that USAP fits the approximate intensity functions estimated by the MCMC better than UNSP.

The better approximation of USAPs in Fig.~\ref{fig:mcmc-vs-vi} comes from the fact that the kernel function can adjust its parameters at each interval independently. While for UNSPs, the CIF at each interval is affected by all the kernel functions that are triggered by the events which happen after that interval and are from the layer immediately below.

In addition, for the CIFs of the approximate posterior point processes at
time \(t\), USAPs can capture the information from both the events
happening before \(t\) and the events happening after \(t\), while UNSPs
can only propagate the information from the future to the past.
Both the events
happening before \(t\) and the events happening after \(t\) have an
influence on the posterior point processes at time \(t\) \citep[Proposition
1]{hong2022deep}. Therefore, USAPs are better than UNSPs, since
UNSPs only include the information of the events happening after \(t\).

\algnewcommand{\linecomment}[1]{
\(\triangleright\) #1}

\begin{algorithm}[H]
\caption{Prediction with MCMC or approximation}
\label{alg:prediction-mcmc}
{\bfseries Input:} observed data \(\mathbf{x}=\{e_1,e_2,\cdots,e_n\}\), where \(e_i=(t_i,k_i)\), sampler \(G(\mathbf{x},\bm{\Theta}^G)\),
and model \(\mathcal{M}\). \\
{\bfseries Initialization:} \(\bm{\Theta}\),  \(\tilde{\bm{\Theta}}\), and sample size \(\mathcal{S}\).\\
{\bfseries Output:} time prediction \(\hat{t}_{n+1}\), type prediction \(\hat{k}_{n+1}\)  \\

\linecomment{If we are predicting using MCMC,

\hspace{2em}\(G(\mathbf{x},\bm{\Theta}^G)=p(\mathbf{z}\mid \mathbf{x};\bm{\Theta})\), where \(\bm{\Theta}^G=\bm{\Theta}\).}

\linecomment{If we are predicting using approximation,

\hspace{2em}\(G(\mathbf{x},\bm{\Theta}^G)=q(\mathbf{z};\mathbf{x},\bm{\Theta}^I)\), where \(\bm{\Theta}^G=\bm{\Theta}^I\).}\\

\begin{algorithmic}[1]
   \For{\(s\) = \(1\) \textbf{to} \(\mathcal{S}\) }
   \State \uline{sample \(\mathbf{z}^s \sim G(\mathbf{x},\bm{\Theta}^G)\)} 
   \EndFor
   \State estimate the CIFs for the top layer based on \(\{\mathbf{z}^s\}_{s=1}^{\mathcal{S}}\) (MLE)\label{alg:prediction-mcmc-update-top}
   \For{\(s\) = \(1\) \textbf{to} \(\mathcal{S}\) }
   \State \uline{sample \(\mathbf{z}^s \sim G(\mathbf{x},\bm{\Theta}^G)\)} 
   \State sample the future time \(\hat{t}^s_{n+1}\) based on \(\mathbf{z}^s\) 
   \State sample the future type \(\hat{k}^s_{n+1}\) based on \(\mathbf{z}^s\)
   \EndFor
   \State \(\hat{t}_{n+1}=\frac{1}{\mathcal{S}}\sum\limits_{s=1}^{\mathcal{S}}\hat{t}^s_{n+1}\)
   \State \(\hat{k}_{n+1}=\argmax\limits_{k \in \{1,\dots,K_0\}}\sum\limits_{s=1}^{\mathcal{S}}\mathds{1}_k(\hat{k}^s_{n+1})\)
\end{algorithmic}
\end{algorithm}

\begin{remark} \label{remark:general-spatio-temporal}
Similar to \citet{hong2022deep}, we can simply replace the kernel function
with a non-causal kernel (\textit{i.e.}, a kernel function that has
positive values for all inputs with any dimensionality) to apply UNSPs in
general Euclidean space (not just a timeline). It is not as
straightforward to apply USAPs to spatial point processes (SPPs) with
multiple dimensions, as we need to identify the ``bounding'' events for any
point in space.
\end{remark}

\eject 
\section{INFERENCE} \label{sec:inference}
\subsection{Inference for the Model Parameters}
While \citet{hong2022deep} view the line \ref{alg:mcem-opt-marginal} in Alg.~\ref{alg:mcem} as a part of ascent-based MCEM, we can also view it
as an unbiased estimator of the gradient of the marginal likelihood \(\log p(\mathbf{x};\mathbf{\Theta})\) based on the Fisher identity \citep{ou2020joint,naesseth2020markovian},
\begin{equation}
    \nabla_{\bm{\Theta}}\log p(\mathbf{x};\bm{\Theta})=\mathbb{E}_{p(\mathbf{z}\mid \mathbf{x};\bm{\Theta})}\left[\nabla_{\bm{\Theta}}\log p(\mathbf{z},\mathbf{x};\bm{\Theta})\right]. \label{eq:unbiased-grad-marginal-likelihood}
\end{equation}
Based on Eq.~\ref{eq:unbiased-grad-marginal-likelihood}, we can update the model parameters by stochastic gradient ascent. The full convergence analysis for the model parameters has not been well-established, and it is still an active research area \citep{neath2013convergence,hong2022deep}.
\subsection{Inference for the Variational Parameters}
Variational inference with the inclusive KL divergence and unbiased gradient has demonstrated the ability to help mitigate the issue of the underestimation of the variance of the posterior \citep{naesseth2020markovian}. The inclusive KL divergence between the true posterior point processes and the approximate point processes \(\infdiv{p}{q^I}\)  is
\begin{equation*}
    \mathbb{E}_{p(\mathbf{z}\mid \mathbf{x};\bm{\Theta})}\left[\log p(\mathbf{z}\mid \mathbf{x};\bm{\Theta}) - \log q(\mathbf{z};\mathbf{x}, \bm{\Theta}^I)\right],
\end{equation*}
where
\(\bm{\Theta}^I=\left\{\left\{\bm{\Theta}^I_{\ell,k}\right\}_{\ell=1}^L\right\}_{k=1}^{K_{\ell}}\),
\(\log q(\mathbf{z};\mathbf{x}, \bm{\Theta}^I)\) is the log-likelihood of the
approximate point processes, and
\begin{align*}
    \log q(\mathbf{z};\mathbf{x}, \bm{\Theta}^I)\!\! &=\!\! \sum_{\ell=1}^L\log q(\mathbf{z}_{\ell};\mathbf{z}_{\ell-1}, \bm{\Theta}^I_{\ell})\\
    &=\!\!\sum_{\ell=1}^L\!\sum_{k=1}^{K_{\ell}}\!\!\left(\! \sum_{t_{\ell,k,j}}\!\!\log \!\lambda^I_{\ell,k}(t_{\ell,k,j})\!\!-\!\!\!\!\int_0^T\!\!\!\!\lambda^I_{\ell,k}(t)dt\!\!\right)\!.
\end{align*}

The gradient of the KL divergence \textit{w.r.t} the variational parameters \(\bm{\Theta}^I\) is 
\begin{equation}
    \mathbb{E}_{p(\mathbf{z} \mid \mathbf{x};\bm{\Theta})}[-\nabla_{\bm{\Theta}^I}\log q(\mathbf{z};\mathbf{x}, \bm{\Theta}^I)], \label{eq:approx-variational-grad}
\end{equation}
because \(p(\mathbf{z} \mid \mathbf{x};\bm{\Theta})\) does not depend on \(\bm{\Theta}^I\).

\citet{naesseth2020markovian} prove the convergence of the variational parameters under some regularity conditions for MCMC with a fixed number of dimensions. However, our Markov chains have an unbounded number of dimensions. We leave the research for the convergence of our variational parameters for future work.

According to Eq.~\ref{eq:approx-variational-grad} and line \ref{alg:mcem-opt-kl} in Alg.~\ref{alg:mcem}, we can see that the gradient of the inclusive KL divergence is identical to the gradient of the log-likelihood of the virtual point processes if we let the virtual point processes be our approximate point processes (\textit{i.e.}, let \(\tilde{\bm{\Theta}}=\bm{\Theta}^I\)). Therefore, we can use Alg.~\ref{alg:mcem} to train our approximate point processes and the parameters for the virtual point processes would just be the parameters for our approximate point processes. Then, when doing sampling for the approximate posterior point processes, we can directly sample from it using the inversion sampling without involving any gradient ascent or MCMC steps. This is especially helpful when doing prediction because we can avoid the time-consuming mixing steps of MCMC for the prediction of each event. We will explain more about this in Sec.~\ref{sec:prediction}.

We use Adam \citep{kingma2014adam} to optimize both the model parameters and variational parameters.
\section{PREDICTION} \label{sec:prediction}

We follow the prediction procedures with MCMC developed by \citet{hong2022deep}
in Alg.~\ref{alg:prediction-mcmc}. Suppose we are given a sequence of events
\(\{e_1,e_2,\cdots,e_n\}\), where \(e_i=(t_i,k_i)\), \(t_i\) and \(k_i\)
represent the time and the type for the \(i\)-th event respectively, and \(t_1
\leq t_2 \leq t_3 \leq \cdots \leq t_n\). We initialize RPPs with parameters
\(\bm{\Theta}\), VPPs with parameters \(\tilde{\bm{\Theta}}\). These parameters
were obtained by MCEM running on the training data. We first generate posterior
samples to estimate the constant rates for the top layer. Suppose the number of
events for the \(s\)-th sample of \(Z_{L,k}\) is \(m^s_{L,k}\) and the time
period is \([0,T]\), then the MLE for the constant rate is
\(\mu_{k}=\frac{1}{\mathcal{S}}\sum_{s=1}^{\mathcal{S}}\frac{m^s_{L,k}}{T}\).
After getting new constant rates for the top layer, we can generate a new set of
posterior samples, starting the initial state of the Markov chain to be the last sample of the previous MCMC sampling. Conditional on the generated posterior sample \(\mathbf{z}\), we can extend the CIFs of RPPs to the future (\textit{i.e.}, the CIFs at the time period \((t_n, \infty)\)), because the CIFs only depend on the history of the events. Then, we can sample the time and the type for the next future event. Suppose the samples for the next future event (\(e_{n+1}\)) are \((t^1_{n+1}, k^1_{n+1}), (t^2_{n+1}, k^2_{n+1}),\dots,(t^\mathcal{S}_{n+1},k^\mathcal{S}_{n+1})\), where \((t^s_{n+1}, k^s_{n+1})\) represents the \(s\)-th sample of the time and the type for the \((n+1)\)-th event, then the time prediction is \(\hat{t}_{n+1}=\frac{1}{\mathcal{S}}\sum_{s=1}^{\mathcal{S}}t^s_{n+1}\) and the type prediction is \(\hat{k}_{n+1}=\argmax_{k \in \{1,\dots,K_0\}}\sum_{s=1}^{\mathcal{S}}\mathds{1}_k(k^s_{n+1})\), where  \(\mathds{1}_k(k_{n+1}^s)\) is equal to 1 \textit{iff} \(k=k_{n+1}^s\). After each prediction, we record the last posterior sample as the initial state of the next MCMC step, and we also update the constant rates for the top layer using MLE.

In Alg.~\ref{alg:prediction-mcmc}, we can replace the MCMC
sampling with the sampling from our approximate posterior point processes to get our proposed method for prediction.
Sampling from \(q(\mathbf{z};\mathbf{x},\bm{\Theta}^I)\) is much faster than
sampling from MCMC, because we can directly use the inversion method
\citep{cinlar2013introduction} to sample instead of performing many MCMC steps. Faster sampling can help us achieve better prediction performance when only a limited amount of time is allowed, which we will use experiments to demonstrate in Sec.~\ref{sec:experiments}.
\newcommand\suffix[1]{#1}
\begin{figure*}[!htbp]
\centering
\includegraphics[width=\textwidth]{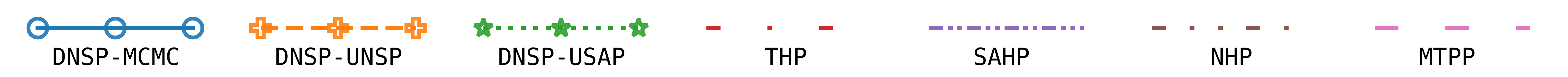}
\end{figure*}
\begin{figure*}[!htbp]
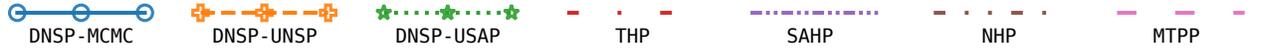
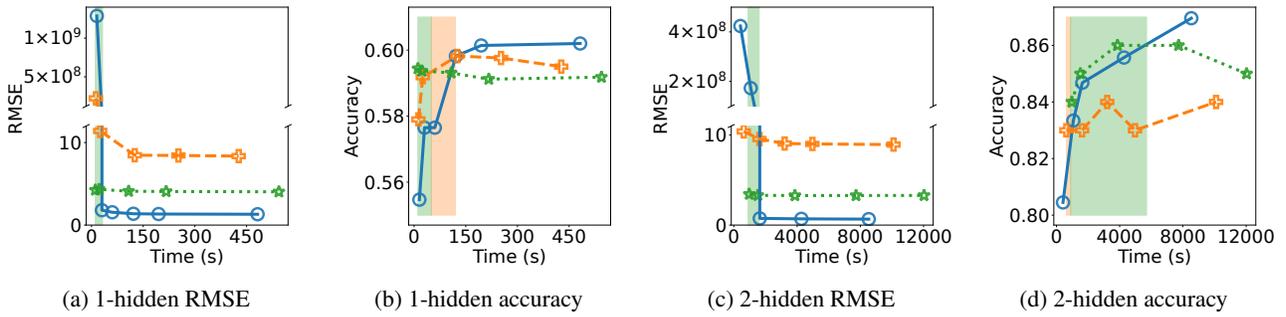

\centering
\begin{subfigure}[b]{.245\textwidth}
         \centering
	    \includegraphics[width=\textwidth]{prediction-accuracy-figures/1-hidden-synthetic-rmse\suffix{.pdf}}
         \caption{1-hidden RMSE}
         \label{fig:synthetic-1-hidden-rmse}
\end{subfigure}
\begin{subfigure}[b]{.245\textwidth}
         \centering
	    \includegraphics[width=\textwidth]{prediction-accuracy-figures/1-hidden-synthetic-accuracy\suffix{.pdf}}
         \caption{1-hidden accuracy}
         \label{fig:synthetic-1-hidden-accuracy}
\end{subfigure}
\begin{subfigure}[b]{.245\textwidth}
         \centering
	    \includegraphics[width=\textwidth]{prediction-accuracy-figures/2-hidden-synthetic-rmse\suffix{.pdf}}
         \caption{2-hidden RMSE}
         \label{fig:synthetic-2-hidden-rmse}
\end{subfigure}
\begin{subfigure}[b]{.245\textwidth}
         \centering
	    \includegraphics[width=\textwidth]{prediction-accuracy-figures/2-hidden-synthetic-accuracy\suffix{.pdf}}
         \caption{2-hidden accuracy}
         \label{fig:synthetic-2-hidden-accuracy}
\end{subfigure}
\caption{Synthetic Dataset}
\label{fig:synthetic}
\end{figure*}
\begin{figure*}[!htbp]
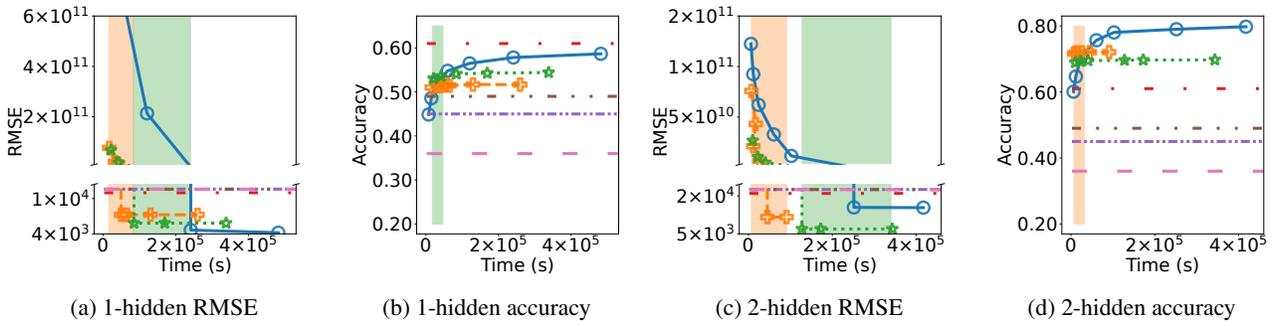

\centering
\begin{subfigure}[b]{.245\textwidth}
         \centering
	    \includegraphics[width=\textwidth]{prediction-accuracy-figures/1-hidden-retweet-rmse\suffix{.pdf}}
         \caption{1-hidden RMSE}
         \label{fig:retweet-1-hidden-rmse}
\end{subfigure}
\begin{subfigure}[b]{.245\textwidth}
         \centering
	    \includegraphics[width=\textwidth]{prediction-accuracy-figures/1-hidden-retweet-accuracy\suffix{.pdf}}
         \caption{1-hidden accuracy}
         \label{fig:retweet-1-hidden-accuracy}
\end{subfigure}
\begin{subfigure}[b]{.245\textwidth}
         \centering
	    \includegraphics[width=\textwidth]{prediction-accuracy-figures/2-hidden-retweet-rmse\suffix{.pdf}}
         \caption{2-hidden RMSE}
         \label{fig:retweet-2-hidden-rmse}
\end{subfigure}
\begin{subfigure}[b]{.245\textwidth}
         \centering
	    \includegraphics[width=\textwidth]{prediction-accuracy-figures/2-hidden-retweet-accuracy\suffix{.pdf}}
         \caption{2-hidden accuracy}
         \label{fig:retweet-2-hidden-accuracy}
\end{subfigure}
\caption{Retweet Dataset}
\label{fig:retweet}
\end{figure*}
\begin{figure*}[!htbp]
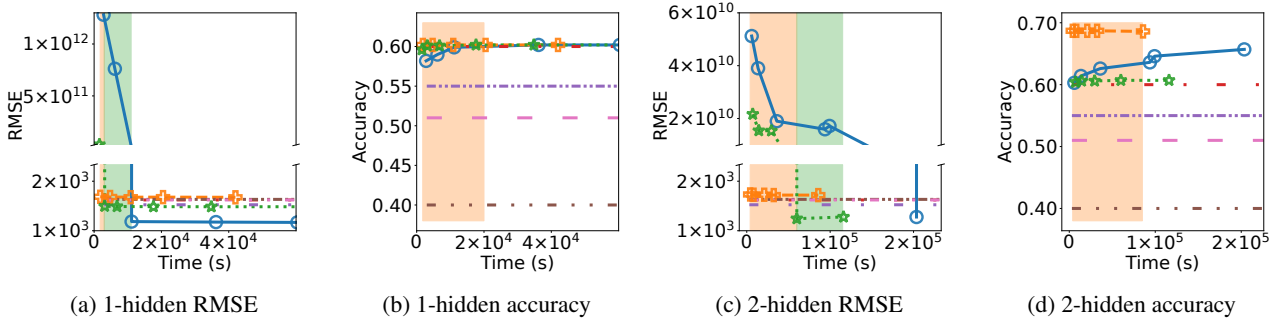

\centering
\begin{subfigure}[b]{.245\textwidth}
         \centering
	    \includegraphics[width=\textwidth]{prediction-accuracy-figures/1-hidden-earthquake-rmse\suffix{.pdf}}
         \caption{1-hidden RMSE}
         \label{fig:earthquake-1-hidden-rmse}
\end{subfigure}
\begin{subfigure}[b]{.245\textwidth}
         \centering
	    \includegraphics[width=\textwidth]{prediction-accuracy-figures/1-hidden-earthquake-accuracy\suffix{.pdf}}
         \caption{1-hidden accuracy}
         \label{fig:earthquake-1-hidden-accuracy}
\end{subfigure}
\begin{subfigure}[b]{.245\textwidth}
         \centering
	    \includegraphics[width=\textwidth]{prediction-accuracy-figures/2-hidden-earthquake-rmse\suffix{.pdf}}
         \caption{2-hidden RMSE}
         \label{fig:earthquake-2-hidden-rmse}
\end{subfigure}
\begin{subfigure}[b]{.245\textwidth}
         \centering
	    \includegraphics[width=\textwidth]{prediction-accuracy-figures/2-hidden-earthquake-accuracy\suffix{.pdf}}
         \caption{2-hidden accuracy}
         \label{fig:earthquake-2-hidden-accuracy}
\end{subfigure}
\caption{Earthquake Dataset}
\label{fig:earthquake}
\end{figure*}
\begin{figure*}[!htbp]
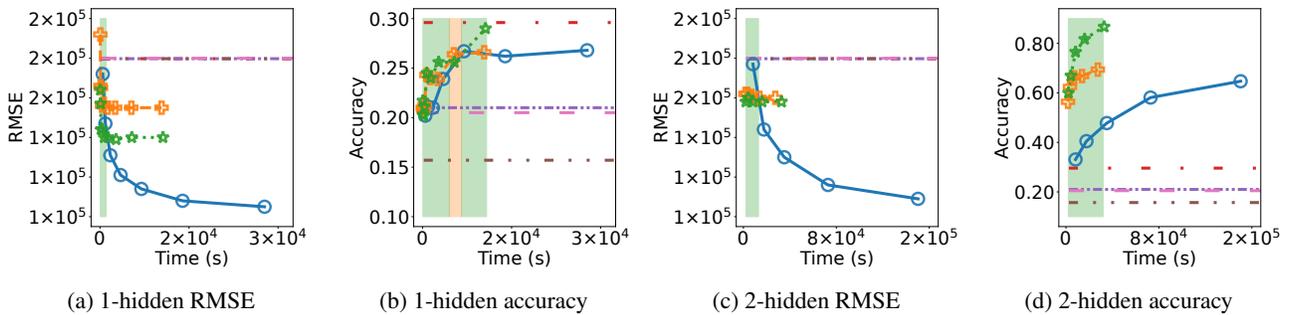

\centering
\begin{subfigure}[b]{.245\textwidth}
         \centering
	    \includegraphics[width=\textwidth]{prediction-accuracy-figures/1-hidden-homicide-rmse\suffix{.pdf}}
         \caption{1-hidden RMSE}
         \label{fig:homicide-1-hidden-rmse}
\end{subfigure}
\begin{subfigure}[b]{.245\textwidth}
         \centering
	    \includegraphics[width=\textwidth]{prediction-accuracy-figures/1-hidden-homicide-accuracy\suffix{.pdf}}
         \caption{1-hidden accuracy}
         \label{fig:homicide-1-hidden-accuracy}
\end{subfigure}
\begin{subfigure}[b]{.245\textwidth}
         \centering
	    \includegraphics[width=\textwidth]{prediction-accuracy-figures/2-hidden-homicide-rmse\suffix{.pdf}}
         \caption{2-hidden RMSE}
         \label{fig:homicide-2-hidden-rmse}
\end{subfigure}
\begin{subfigure}[b]{.245\textwidth}
         \centering
	    \includegraphics[width=\textwidth]{prediction-accuracy-figures/2-hidden-homicide-accuracy\suffix{.pdf}}
         \caption{2-hidden accuracy}
         \label{fig:homicide-2-hidden-accuracy}
\end{subfigure}
\caption{Homicide Dataset}
\label{fig:homicide}
\end{figure*}

\section{EXPERIMENTS} \label{sec:experiments}
The code is available online at \url{https://github.com/hongchengkuan/Inclusive_VI_NSPs}. We implement all the algorithms on the same code base using Pytorch, so the implementation has little influence on the time comparison. 

Following \citet{hong2022deep}, we use root mean square error (RMSE) to compare
the time prediction and use accuracy to compare the type prediction. The models
we use for DNSPs are 1-hidden and 2-hidden as explained in \citet{hong2022deep}.
Each dataset is split into training, validation, and test sets. We use training
sets to train our model parameters and variational parameters, use validation
sets to stop early, and use test sets to calculate the metrics used to compare
the performance. When performing prediction, we use different numbers of samples
to measure the performance. Different numbers of samples correspond to different
computational time budgets, as more time is needed for a larger sample size. We add a tiny background base rate \((1\times 10^{-10})\) to the intensity for the observation to prevent NaN and Inf issues. More details about the experiments can be found in Appx.~\ref{sec:experiments-details}.

In Figs.~\ref{fig:synthetic}, \ref{fig:retweet}, \ref{fig:earthquake} and
\ref{fig:homicide}, we use different line styles with different colors to
represent the results from different models or algorithms, where DNSP-MCMC represents the results obtained by performing prediction using MCMC for DNSPs, DNSP-UNSP represents the results obtained by performing prediction using approximation with UNSPs for DNSPs, DNSP-USAP represents the results obtained by performing prediction using approximation with USAPs, THP represents transformer Hawkes process
\citep{zuo2020transformer}, SAHP represents self-attentive Hawkes process 
\citep{zhang2020self}, NHP represents neural Hawkes process  \citep{mei2017neural}, and MTPP is a multitask point process model
\citep{lian2015multitask}. The vertical axes represent
the RMSE or accuracy. The horizontal axes represent the time used to do sampling with various sample sizes for the predictions for all events from all sequences. Among UNSPs, USAPs, and MCMC, UNSPs are the best in the areas filled with light orange, and USAPs are the best in the areas filled with light green. A table of results with performance compared at fixed time-points is presented in Appx.~\ref{appx:experiments-table}, where "/" means there is no result for that entry.

Our experiments show that when only a limited amount of time is available, UNSPs
and USAPs perform better than MCMC for both time prediction and type prediction.
USAPs are clearly better than UNSPs in terms of time prediction, and
the type predictions of USAPs and UNSPs are similar, although USAPs require more computational resources than UNSPs. Moreover, DNSPs-based methods
are better than non-DNSPs-based methods for all these real-world datasets.

\subsection{Synthetic Data Experiments}
We construct 2 synthetic datasets. One dataset is generated from a 1-hidden model
and the other is generated from a 2-hidden model. We use the 1-hidden DNSP to learn
from the dataset generated from 1-hidden, and the 2-hidden DNSP to learn from the dataset generated from 2-hidden. For simplicity, \(k\) is set to be 1 and fixed for the Weibull kernel, so that it becomes an exponential function.

Fig.~\ref{fig:synthetic} summarizes the results for our synthetic datasets.  
It shows that when we increase the sample size, the prediction performance of
UNSPs, USAPs, and MCMC become better, at a cost of running time. The
improvement of MCMC is more significant, as MCMC becomes closer to the true posterior point processes when we sample more from MCMC, while USAPs and UNSPs will never converge to the true posterior point processes. It also demonstrates that USAPs and UNSPs can achieve better prediction results for both the time and the type than MCMC when only a small period of time is allowed. When we can run our programs long enough, MCMC achieves the best results for all these experiments, which is reasonable because MCMC converges to the true posterior distribution when the sample size goes to infinity and the synthetic data is generated from DNSPs.

USAPs are better than UNSPs with regards to RMSE for both 1-hidden and 2-hidden datasets (Figs.~\ref{fig:synthetic-1-hidden-rmse} and \ref{fig:synthetic-2-hidden-rmse}). There is a significant drop for the RMSE when we increase the sample size; it is because we may have no or very few hidden events in our posterior samples when the sample size is too small, causing the CIFs for the top layer or for the future to be very small. With a small CIF, the sample for the next future event would be very far away from the next true future event.

For accuracy, USAPs are clearly better than UNSPs for the 2-hidden synthetic dataset (Fig.~\ref{fig:synthetic-2-hidden-accuracy}) and they have similar performance for the 1-hidden synthetic dataset (Fig.~\ref{fig:synthetic-1-hidden-accuracy}). 
\subsection{Real-World Data Experiments}  \label{sec:real-world-experiments}
Similar to \citet{hong2022deep}, the datasets we use are of retweets \citep{zhao2015seismic},
earthquakes \citep{ncedc,bdsn,hrsn,brad}, and homicides \citep{chicagocrime}. More details of the datasets can be found in
Appx.~\ref{sec:experiments-details}. We compare MCMC, UNSPs, and USAPs for DNSPs
with other state-of-the-art methods: transformer Hawkes process (THP)
\citep{zuo2020transformer}, self-attentive Hawkes process (SAHP)
\citep{zhang2020self}, neural Hawkes process (NHP) \citep{mei2017neural}, and MTPP
\citep{lian2015multitask}. THP, SAHP, and NHP are neural-network-based Hawkes
processes. MTPP is a Cox-process-based model. We split each real-world dataset
into training, validation, and test sets. We train, validate and test for all
these models using the same split. For retweets, we use mini-batch gradient
ascent, and we use batch gradient ascent for other datasets. The training and
prediction procedures of the methods are the same as in \citet{hong2022deep}. We
do not fix \(k\) for the Weibull kernel as we did in the synthetic data experiments.

Figs.~\ref{fig:retweet}, \ref{fig:earthquake} and \ref{fig:homicide} summarize the experimental results for retweets, earthquakes, and homicides respectively. Similar to the experimental results for the synthetic dataset, UNSPs and USAPs are better than MCMC for both the time prediction and the type prediction when we only have a small number of samples from the approximate posterior distribution. As we increase the sample size, UNSPs, USAPs, and MCMC all gain some improvement to various degrees.

For the time prediction, DNSPs with MCMC or approximate posterior point processes achieve the best prediction results in all these experiments (Figs.~\ref{fig:retweet-1-hidden-rmse}, \ref{fig:retweet-2-hidden-rmse}, \ref{fig:earthquake-1-hidden-rmse}, \ref{fig:earthquake-2-hidden-rmse},  \ref{fig:homicide-1-hidden-rmse}, and \ref{fig:homicide-2-hidden-rmse}). UNSPs or USAPs are better than other non-DNSPs baselines for all these experiments (Figs.~\ref{fig:retweet-1-hidden-rmse}, \ref{fig:retweet-2-hidden-rmse}, \ref{fig:earthquake-1-hidden-rmse}, \ref{fig:earthquake-2-hidden-rmse},  \ref{fig:homicide-1-hidden-rmse}, and \ref{fig:homicide-2-hidden-rmse}). USAPs are better than UNSPs for all of these experiments in the end (Figs.~\ref{fig:retweet-1-hidden-rmse}, \ref{fig:retweet-2-hidden-rmse}, \ref{fig:earthquake-1-hidden-rmse}, \ref{fig:earthquake-2-hidden-rmse},  \ref{fig:homicide-1-hidden-rmse}, and \ref{fig:homicide-2-hidden-rmse}). UNSPs are better than USAPs for retweets and earthquakes when we only have a small number of samples (Figs.~\ref{fig:retweet-1-hidden-rmse}, \ref{fig:retweet-2-hidden-rmse}, \ref{fig:earthquake-1-hidden-rmse}, and \ref{fig:earthquake-2-hidden-rmse}). USAPs are always better than UNSPs for homicides in terms of time prediction (Figs.~\ref{fig:homicide-1-hidden-rmse} and \ref{fig:homicide-2-hidden-rmse}). 

For the type prediction, DNSPs with MCMC or approximate posterior point processes achieve the best prediction results for all these datasets (Fig.~\ref{fig:retweet-2-hidden-accuracy} for retweets, \ref{fig:earthquake-2-hidden-accuracy} for earthquakes, Fig.~\ref{fig:homicide-2-hidden-accuracy} for homicides). UNSPs or USAPs are better than other non-DNSPs baselines for all these datasets (Fig.~\ref{fig:retweet-2-hidden-accuracy} for retweets, \ref{fig:earthquake-2-hidden-accuracy} for earthquakes, Fig.~\ref{fig:homicide-2-hidden-accuracy} for homicides). In Figs.~\ref{fig:retweet-1-hidden-accuracy} and \ref{fig:homicide-2-hidden-accuracy}, USAPs are better than UNSPs, while UNSPs are better than USAPs in Figs.~\ref{fig:retweet-2-hidden-accuracy} and  \ref{fig:earthquake-2-hidden-accuracy}. USAPs and UNSPs have similar performance in Figs.~\ref{fig:earthquake-1-hidden-accuracy} and \ref{fig:homicide-1-hidden-accuracy}.

Compared with the results in \citet{hong2022deep}, DNSP-MCMC in this paper is better in terms of accuracy and RMSE. The performance of MCMC differs because we use a Weibull kernel in this paper, while \citet{hong2022deep} adopt a gamma kernel, as explained in Sec.~\ref{sec:kernel-function}.
\subsection{Computational Complexity}
The full analysis requires the bound for the mixing steps of MCMC. The analysis is not trivial and we do not have this bound \cite[see][Appx.~G.5]{hong2022deep}.
\section{CONCLUSION} \label{sec:conclusion}
VI and point processes have both attracted great attention due to their excellent
performance. However, little attention has been given to developing VI algorithms
for point processes with hierarchical structures. We develop the first VI
algorithm for NSPs (a class of point processes with hierarchical structures) in
this paper.

Typically, posterior inference for NSPs is considered a very hard problem, and MCMC is required, as it involves an unbounded number of points in the posterior point processes. We incorporate MCMC into our VI algorithm, treating the samples from our approximate posterior point processes as the candidates for the posterior point processes. During training processes, we gradually update our approximations to make our approximations become closer and closer to the posterior point processes through the minimization of the inclusive KL divergence.  Our experiments show that our approximate posterior point processes (USAPs and UNSPs) can fit the true posterior point processes very well. When time constraint is a concern, USAPs and UNSPs provide a very good alternative to MCMC.

In our VI algorithm, we bring 4 active research areas (MCMC, VI, neural
networks, and point processes) together, and we have found many research topics that we believe will be of interest to many researchers in these areas, \textit{e.g.}, analysis of the mixing time of MCMC, convergence analysis of variational parameters, efficient and well-behaved architectures of neural networks, and the construction of spatio-temporal point processes with hierarchies.





\subsubsection*{Acknowledgements}
Computational resources were provided through Office of Naval Research grant N00014-18-1-2252.

Data for this study came from
the Berkeley Digital Seismic Network (BDSN), doi:10/7932/BDSN;
the High Resolution Seismic Network (HRSN), doi:10.7932/HRSN; and
the Bay Area Regional Deformation Network (BARD), doi:10.7932/BARD,
all operated by the UC Berkeley Seismological Laboratory and
archived at the Northern California Earthquake Data center (NCEDC), doi: 10/7932/NCEDC.


\bibliography{sample_paper}

\begin{thebibliography}{}

\bibitem[Ba et~al., 2016]{ba2016layer}
Ba, J.~L., Kiros, J.~R., and Hinton, G.~E. (2016).
\newblock Layer normalization.

\bibitem[BARD, 2014]{brad}
BARD (2014).
\newblock Bay area regional deformation network. {UC Berkeley Seismological
  Laboratory}. dataset.

\bibitem[BDSN, 2014]{bdsn}
BDSN (2014).
\newblock Berkeley digital seismic network. {UC Berkeley Seismological
  Laboratory}. dataset.

\bibitem[Blei et~al., 2017]{blei2017variational}
Blei, D.~M., Kucukelbir, A., and McAuliffe, J.~D. (2017).
\newblock Variational inference: A review for statisticians.
\newblock {\em Journal of the American statistical Association},
  112(518):859--877.

\bibitem[\c{C}inlar, 2013]{cinlar2013introduction}
\c{C}inlar, E. (2013).
\newblock {\em Introduction to stochastic processes}.
\newblock Dover Publications.

\bibitem[Chen et~al., 2021]{chen2021neural}
Chen, R. T.~Q., Amos, B., and Nickel, M. (2021).
\newblock Neural spatio-temporal point processes.
\newblock In {\em International Conference on Learning Representations}.

\bibitem[COC, 2022]{chicagocrime}
COC (2022).
\newblock City of {C}hicago, {C}rimes - 2001 to present.
\newblock
  \url{https://data.cityofchicago.org/Public-Safety/Crimes-2001-to-Present/ijzp-q8t2}.
\newblock Accessed: 2022-08-14.

\bibitem[Hendrycks and Gimpel, 2016]{hendrycks2016gaussian}
Hendrycks, D. and Gimpel, K. (2016).
\newblock Gaussian error linear units ({GELU}s).

\bibitem[Hong and Shelton, 2022]{hong2022deep}
Hong, C. and Shelton, C. (2022).
\newblock Deep {N}eyman-{S}cott processes.
\newblock In Camps-Valls, G., Ruiz, F. J.~R., and Valera, I., editors, {\em
  Proceedings of The 25th International Conference on Artificial Intelligence
  and Statistics}, volume 151 of {\em Proceedings of Machine Learning
  Research}, pages 3627--3646. PMLR.

\bibitem[HRSN, 2014]{hrsn}
HRSN (2014).
\newblock High resolution seismic network. {UC Berkeley Seismological
  Laboratory}. dataset.

\bibitem[Jordan et~al., 1999]{jordan1999introduction}
Jordan, M.~I., Ghahramani, Z., Jaakkola, T.~S., and Saul, L.~K. (1999).
\newblock An introduction to variational methods for graphical models.
\newblock {\em Machine learning}, 37(2):183--233.

\bibitem[Kingma and Ba, 2015]{kingma2014adam}
Kingma, D.~P. and Ba, J. (2015).
\newblock Adam: {A} method for stochastic optimization.
\newblock In Bengio, Y. and LeCun, Y., editors, {\em 3rd International
  Conference on Learning Representations, {ICLR} 2015, San Diego, CA, USA, May
  7-9, 2015, Conference Track Proceedings}.

\bibitem[Kingma and Welling, 2014]{kingma2014auto}
Kingma, D.~P. and Welling, M. (2014).
\newblock Auto-encoding variational bayes.
\newblock In Bengio, Y. and LeCun, Y., editors, {\em 2nd International
  Conference on Learning Representations, {ICLR} 2014, Banff, AB, Canada, April
  14-16, 2014, Conference Track Proceedings}.

\bibitem[Lian et~al., 2015]{lian2015multitask}
Lian, W., Henao, R., Rao, V., Lucas, J., and Carin, L. (2015).
\newblock A multitask point process predictive model.
\newblock In Bach, F. and Blei, D., editors, {\em Proceedings of the 32nd
  International Conference on Machine Learning}, volume~37 of {\em Proceedings
  of Machine Learning Research}, pages 2030--2038, Lille, France. PMLR.

\bibitem[Linderman et~al., 2017]{linderman2017bayesian}
Linderman, S.~W., Wang, Y., and Blei, D.~M. (2017).
\newblock Bayesian inference for latent {H}awkes processes.
\newblock In {\em Advances in Approximate Bayesian Inference}.

\bibitem[Mei and Eisner, 2017]{mei2017neural}
Mei, H. and Eisner, J.~M. (2017).
\newblock The neural {H}awkes process: A neurally self-modulating multivariate
  point process.
\newblock In Guyon, I., Luxburg, U.~V., Bengio, S., Wallach, H., Fergus, R.,
  Vishwanathan, S., and Garnett, R., editors, {\em Advances in Neural
  Information Processing Systems}, volume~30. Curran Associates, Inc.

\bibitem[M{\o}ller and Waagepetersen, 2003]{moller2003statistical}
M{\o}ller, J. and Waagepetersen, R.~P. (2003).
\newblock {\em Statistical inference and simulation for spatial point
  processes}.
\newblock CRC Press.

\bibitem[Naesseth et~al., 2020]{naesseth2020markovian}
Naesseth, C., Lindsten, F., and Blei, D. (2020).
\newblock Markovian score climbing: Variational inference with {KL}(p \(\vert
  \vert\) q).
\newblock In Larochelle, H., Ranzato, M., Hadsell, R., Balcan, M., and Lin, H.,
  editors, {\em Advances in Neural Information Processing Systems}, volume~33,
  pages 15499--15510. Curran Associates, Inc.

\bibitem[NCEDC, 2014]{ncedc}
NCEDC (2014).
\newblock Northern {C}alifornia earthquake data center. {UC Berkeley
  Seismological Laboratory}. dataset.

\bibitem[Neath et~al., 2013]{neath2013convergence}
Neath, R.~C. et~al. (2013).
\newblock On convergence properties of the {M}onte {C}arlo {EM} algorithm.
\newblock In {\em Advances in modern statistical theory and applications: a
  Festschrift in Honor of Morris L. Eaton}, pages 43--62. Institute of
  Mathematical Statistics.

\bibitem[Neyman and Scott, 1958]{neyman1958statistical}
Neyman, J. and Scott, E.~L. (1958).
\newblock Statistical approach to problems of cosmology.
\newblock {\em Journal of the Royal Statistical Society: Series B
  (Methodological)}, 20(1):1--29.

\bibitem[Ou and Song, 2020]{ou2020joint}
Ou, Z. and Song, Y. (2020).
\newblock Joint stochastic approximation and its application to learning
  discrete latent variable models.
\newblock In Peters, J. and Sontag, D., editors, {\em Proceedings of the 36th
  Conference on Uncertainty in Artificial Intelligence (UAI)}, volume 124 of
  {\em Proceedings of Machine Learning Research}, pages 929--938. PMLR.

\bibitem[Park et~al., 2022]{park2022an}
Park, J., Chang, W., and Choi, B. (2022).
\newblock An interaction {N}eyman–{S}cott point process model for coronavirus
  disease-19.
\newblock {\em Spatial Statistics}, 47:100561.

\bibitem[Ranganath et~al., 2014]{ranganath2014black}
Ranganath, R., Gerrish, S., and Blei, D. (2014).
\newblock Black box variational inference.
\newblock In Kaski, S. and Corander, J., editors, {\em Proceedings of the
  Seventeenth International Conference on Artificial Intelligence and
  Statistics}, volume~33 of {\em Proceedings of Machine Learning Research},
  pages 814--822, Reykjavik, Iceland. PMLR.

\bibitem[Vaswani et~al., 2017]{vaswani2017attention}
Vaswani, A., Shazeer, N., Parmar, N., Uszkoreit, J., Jones, L., Gomez, A.~N.,
  Kaiser, {\L}., and Polosukhin, I. (2017).
\newblock Attention is all you need.
\newblock In Guyon, I., Luxburg, U.~V., Bengio, S., Wallach, H., Fergus, R.,
  Vishwanathan, S., and Garnett, R., editors, {\em Advances in Neural
  Information Processing Systems}, volume~30. Curran Associates, Inc.

\bibitem[Williams et~al., 2020]{williams2020point}
Williams, A., Degleris, A., Wang, Y., and Linderman, S. (2020).
\newblock Point process models for sequence detection in high-dimensional
  neural spike trains.
\newblock In Larochelle, H., Ranzato, M., Hadsell, R., Balcan, M., and Lin, H.,
  editors, {\em Advances in Neural Information Processing Systems}, volume~33,
  pages 14350--14361. Curran Associates, Inc.

\bibitem[Zhang et~al., 2020]{zhang2020self}
Zhang, Q., Lipani, A., Kirnap, O., and Yilmaz, E. (2020).
\newblock Self-attentive {H}awkes process.
\newblock In III, H.~D. and Singh, A., editors, {\em Proceedings of the 37th
  International Conference on Machine Learning}, volume 119 of {\em Proceedings
  of Machine Learning Research}, pages 11183--11193. PMLR.

\bibitem[Zhao et~al., 2015]{zhao2015seismic}
Zhao, Q., Erdogdu, M.~A., He, H.~Y., Rajaraman, A., and Leskovec, J. (2015).
\newblock Seismic: A self-exciting point process model for predicting tweet
  popularity.
\newblock In {\em Proceedings of the 21th ACM SIGKDD international conference
  on knowledge discovery and data mining}, pages 1513--1522.

\bibitem[Zuo et~al., 2020]{zuo2020transformer}
Zuo, S., Jiang, H., Li, Z., Zhao, T., and Zha, H. (2020).
\newblock Transformer {H}awkes process.
\newblock In III, H.~D. and Singh, A., editors, {\em Proceedings of the 37th
  International Conference on Machine Learning}, volume 119 of {\em Proceedings
  of Machine Learning Research}, pages 11692--11702. PMLR.

\end{thebibliography}




\appendix
\onecolumn

\section{NETWORK STRUCTURES FOR USAPS} \label{sec:self-attention-details}
\paragraph{Event Embedding} Each event \(e_{\ell,i,j}=(t_{\ell,i,j},\text{ppid}_{\ell,i}=\sum_{w=0}^{\ell-1}K_w+i)\) consists of the time \(t_{\ell,i,j}\) and the identifier \(\text{ppid}_{\ell,i}\). Similar to \citet{vaswani2017attention,zuo2020transformer,zhang2020self}, we first encode the event time into a \(d_M\)-dimensional vector \(\bm{pe}_{\ell,i,j}\) though positional encoding,
\begin{equation*}
    \bm{pe}^k_{\ell,i,j}=\begin{cases}
\cos(t_{\ell,i,j}/10000^{\frac{k-1}{d_M}})\text{, if \(k\) is odd,}\\ \sin(t_{\ell,i,j}/10000^{\frac{k}{d_M}})\text{, if \(k\) is even,}
    \end{cases}
\end{equation*}
where \(\bm{pe}^k_{\ell,i,j}\) is the \(k\)-th dimension of \(\bm{pe}_{\ell,i,j}\).

We also train an embedding matrix \(W_{eid} \in \mathbb{R}^{d_M \times n_{pp}}\), where \(n_{pp}=\sum_{w=0}^LK_w\), for the identifiers. For each event with identifier \(\text{ppid}_{\ell,i}\), the embedding for the identifier is \(W_{eid}\cdot \mathbf{p}_{\ell,i}\), where \(\mathbf{p}_{\ell,i} \in \mathbb{R}^{n_{pp} \times 1}\) is a one-hot vector. The \((\sum_{w=0}^{\ell-1}K_w+i)\)-th entry of \(\mathbf{p}_{\ell,i}\) is 1, and the other entries are all 0.

To incorporate the information from both the time and the identifier, the embedding \(x^e_{\ell,i,j}\) of each event \(e_{\ell,i,j}\) can be represented as 
\begin{equation*}
    \bm{x}^e_{\ell,i,j} = \bm{pe}_{\ell,i,j} + W_{eid}\cdot \mathbf{p}_{\ell,i}.
\end{equation*}
\paragraph{Self-Attention} The parameters of the intensity function \(\lambda^I_{\ell,k}(t)\) are determined by all the events from the point processes which are connected to the point process \(Z_{\ell-1,i}\). For each interval \((t_{\ell-1,i,j-1}, t_{\ell-1,i,j}]\), we use self-attention \citep{vaswani2017attention,zhang2020self,zuo2020transformer} to encode the events in the layer immediately below to a hidden vector, 
\begin{align}
    \bm{h}^a_{\ell-1,i,j}=&\left(\sum_{t=1}^{m_{\ell-1,i}}f\left(W_q\cdot \text{LayerNorm}(\bm{x}^e_{\ell-1,i,j}), W_k\cdot \bm{x}^e_{\ell-1,i,t}\right)\cdot \left(W_v\cdot \bm{x}^e_{\ell-1,i,t}\right)\right) \nonumber\\
    &/\sum_{t=1}^{m_{\ell-1,i}}f\left(W_q\cdot \text{LayerNorm}(\bm{x}^e_{\ell-1,i,j}), W_k\cdot \bm{x}^e_{\ell-1,i,t}\right), \label{eq:single-head-attention}
\end{align}
where LayerNorm(\(\cdot\)) is a layer normalization \citep{ba2016layer} operation, \(W_q \in \mathbb{R}^{d_k \times d_M}\), \(W_k \in \mathbb{R}^{d_k \times d_M}\), and \(W_v \in \mathbb{R}^{d_v \times d_M}\) are all linear transformation matrices that transfer the event embedding vectors to queries, keys and values respectively, and \(f(\bm{x}_1, \bm{x}_2)=\exp(\bm{x}_1^T\bm{x}_2)\) is a similarity function to capture the relationship between a query and a key. Notice that unlike the self-attention adopted in \citep{zhang2020self,zuo2020transformer}, we not only consider the influence of the events that happened before the query, but the events that happened after the query. Because for each hidden point process, the posterior distribution of the hidden events can be influenced by all the events from the point processes that are connected to this hidden point process \citep[Proposition
1]{hong2022deep}.

Eq.~\ref{eq:single-head-attention} works as a single-head self-attention, we can also concatenate multiple single-head attentions together to build a multi-head attention \citep{vaswani2017attention}.
We can rewrite Eq.~\ref{eq:single-head-attention} as 
\begin{equation*}
     \bm{h}^a_{\ell-1,i,j}=\text{Self-Attention}(\bm{x}^e_{\ell-1,i,j}, W_q, W_k, W_v),
\end{equation*}
then the multi-head attention version for \(\bm{h}^a_{\ell-1,i,j}\) would be 
\begin{align}
     \bm{h}^a_{\ell-1,i,j}=&\text{Multi-Head-Self-Attention}(\bm{x}^e_{\ell-1,i,j})\nonumber \\
     =&\text{Concat}(\text{head}_1, \cdots, \text{head}_h)\cdot W_o, \label{eq:multihead-self-attention}
\end{align}
where \(\text{head}_d = \text{Self-Attention}(\bm{x}^e_{\ell-1,i,j}, W^d_q, W^d_k, W^d_v)\), \(W^d_q \in \mathbb{R}^{d_k \times d_M}\), \(W^d_k \in \mathbb{R}^{d_k \times d_M}\), \(W^d_v \in \mathbb{R}^{d_v \times d_M}\), and \(W_o \in \mathbb{R}^{hd_v \times d_M}\).

Multi-head self-attention can also be stacked into a deep structure, but we do not have this deep structure in our experiment.
\paragraph{Position-Wise Feed-Forward Layer}
The hidden vector \(\bm{h}^a_{\ell-1,i,j}\) and the residual \(\bm{x}^e_{\ell-1,i,j}\) are then fed into a position-wise feed-forward layer with the number of neurons in the hidden layer as \(d_H\), generating the final output for the hidden vector, 
\begin{equation}
    \bm{h}_{\ell-1,i,j}=\left(W^F_2(\text{GELU}(W^F_1 (\text{LayerNorm}(\text{input})) + \bm{b}_1^F))+\bm{b}_2^F\right)+\text{input}, \label{eq:point-wise-feed-forward}
\end{equation}
where \(\text{input}=\bm{h}^a_{\ell-1,i,j}+\bm{x}^e_{\ell-1,i,j}\), \(W^F_1 \in \mathbb{R}^{d_H \times d_M}\), \(\bm{b}^F_1 \in \mathbb{R}^{d_H \times 1}\), \(W^F_2 \in \mathbb{R}^{d_M \times d_H}\), \(\bm{b}^F_2 \in \mathbb{R}^{d_M \times 1}\) , and GELU is the Gaussian Error Linear Unit \citep{hendrycks2016gaussian}.
\paragraph{Output for the Parameters} We apply a linear transformation to the final hidden vector \(h_{\ell-1,i,j}\) to get the kernel parameters, 
\begin{equation*}
    \bm{\theta}^{I}_{\ell,k}=W^{\theta}_{\ell,k}h_{\ell-1,i,j}+\bm{b}^{\theta}_{\ell,k},
\end{equation*}
where \(W^{\theta}_{\ell,k}\in \mathbb{R}^{n_{\theta} \times d_M}\), \(\bm{b}^{\theta}_{\ell,k} \in \mathbb{R}^{n_{\theta} \times 1}\), and \(n_{\theta}\) is the number of parameters for \(\bm{\theta}^{I}_{\ell,k}\).
\section{EXPERIMENT DETAILS} \label{sec:experiments-details}
\subsection{Datasets}
\paragraph{Synthetic Datasets} The point processes are defined in the interval \((0,20]\). The constant rate for the top layer is set to be 0.15. We fix the kernel parameters and \(k\) is set to be 1 for the Weibull kernel functions. The sampling procedure is the same as the generative process, and we do this for both the 1-hidden and 2-hidden models.
\paragraph{Retweets} \citep{zhao2015seismic} The retweet dataset consists
of a set of tweet streams. The original tweet of a tweet stream has time 0, and the times of the other tweets are the retweet times relative to the original tweet. We have 3 types of events (small, medium, and large) for different numbers of followers. The "small" group consists of users with fewer than 120 users, the "medium" group consists of users with greater than 120 users but fewer than 1363 users, and the "large" group consists of the rest of the users.

\paragraph{Earthquakes} \citep{ncedc,bdsn,hrsn,brad} The earthquake dataset contains the times and magnitudes of the earthquakes collected from northern California earthquake catalog. The times of the earthquakes are constrained to be from 01/01/2014 00:00:00 to 01/01/2020 00:00:00. The region is spanning between \(34.5^{\circ}\) and \(43.2^{\circ}\) latitude and between \(-126.00^{\circ}\) and \(-117.76^{\circ}\) longitude. We divide the earthquakes into 2 types. The earthquakes with a magnitude smaller than 1 are small earthquakes, and the others are big earthquakes.

\paragraph{Homicides} \citep{chicagocrime} The homicide dataset includes the information of homicides that happened in Chicago. We choose 5 contiguous regions in Chicago with the most homicides. The time for the homicides is from 01/01/2001 00:00:00 to 01/01/2020
00:00:00. 5 types of this dataset correspond to 5 different regions in Chicago. The terms of use can be found in  \url{https://www.chicago.gov/city/en/narr/foia/data_disclaimer.html}.

The statistics of the datasets are shown in Table \ref{table:dataset-stats}.
\begin{table}[H]
\caption{Datasets Statistics}
\label{table:dataset-stats}
\centering
\begin{small}
\begin{sc}
\begin{tabular}{lcrrrr}
\toprule
\multirow{2}{*}{Datasets} & \multirow{2}{*}{\# of types} & \multirow{2}{*}{\#  of  predictions} & \multicolumn{3}{c}{\# of sequences}\\
\cmidrule{4-6}
&  & & training &validation &test \\
\midrule
1-hidden Synthetic & 2 & 1563 & 1000 & 100 & 100\\
2-hidden Synthetic & 2 & 4664 & 1000 & 100 & 100\\
Retweets &3    &216465 &20000 & 2000 & 2000 \\
Earthquakes &2    & 25646  & 209 & 53 & 53 \\
Homicides &5    & 997 & 6 & 2 & 2 \\
\bottomrule
\end{tabular}
\end{sc}
\end{small}
\end{table}
\subsection{Self-Attention Training Details}
For all our experiments, we use 1 layer of multi-head self-attention block with 4 heads. The number of dimensions can be found in Table \ref{table:self-attention-dimensions}, where \(d_k\), \(d_v\), \(d_M\), and \(d_H\) appear in Eqs.~\ref{eq:multihead-self-attention} and \ref{eq:point-wise-feed-forward}, MPSS stands for model parameters step size, and SASS stands for self-attention step size. 
\begin{table}[H]
\caption{Self-Attention Dimensions}
\label{table:self-attention-dimensions}
\centering
\begin{small}
\begin{sc}
\begin{tabular}{lcrrrrrr}
\toprule
Datasets & Model & \(d_k\) & \(d_v\) & \(d_M\) & \(d_H\) & MPSS & SASS \\
\midrule
1-hidden Synthetic & USAP(1-hidden) & 8 & 8 & 32 & 64 & 0.01 & 0.010\\
\midrule
2-hidden Synthetic & USAP(2-hidden) & 8 & 8 & 32 & 64 & 0.01 & 0.010\\
\midrule
\multirow{2}{*}{Retweets} &USAP(1-hidden)    &16 &16 & 64 & 128 & 1.00 & 0.001\\
&USAP(2-hidden) &16 &16 & 64 & 128 & 1.00 & 0.001\\
\midrule
\multirow{2}{*}{Earthquakes} &USAP(1-hidden)    &16 &16 & 64 & 128 & 1.00 & 0.001 \\
&USAP(2-hidden) &16 &16 & 64 & 128 & 1.00 & 0.010\\
\midrule
\multirow{2}{*}{Homicides} &USAP(1-hidden)    & 8 & 8 & 32 & 64 & 1.00 & 0.010\\
&USAP(2-hidden) &16 &16 & 64 & 128 & 1.00 & 0.010\\
\bottomrule
\end{tabular}
\end{sc}
\end{small}
\end{table}
\subsection{Experiments} \label{appx:experiments-table}
\begin{table}[H]
\caption{1-Hidden RMSE For Synthetic Dataset}
\label{table:1-hidden_rmse_synthetic}
\centering
\begin{small}
\begin{sc}
\begin{tabular}{cccccc}
\toprule
 \multirow{2}{*}{Model} & \multicolumn{5}{c}{Time(s)} \\ \cmidrule{2-6} & 16 & 32 & 64 & 128 & 256 \\
\midrule
 DNSP-MCMC & 1.24e+09 & 1.79e+00 & 1.54e+00 & 1.37e+00 & 1.33e+00 \\
\midrule
 DNSP-UNSP & 1.63e+08 & 1.12e+01 & 1.03e+01 & 8.46e+00 & 8.43e+00\\
 \midrule
 DNSP-USAP & 4.28e+00 & 4.30e+00 & 4.21e+00 & 4.09e+00 & 4.06e+00 \\
\bottomrule
\end{tabular}
\end{sc}
\end{small}
\end{table}

\begin{table}[H]
\caption{1-Hidden Accuracy For Synthetic Dataset}
\label{table:1-hidden_accuracy_synthetic}
\centering
\begin{small}
\begin{sc}
\begin{tabular}{cccccc}
\toprule
 \multirow{2}{*}{Model} & \multicolumn{5}{c}{Time(s)} \\ \cmidrule{2-6} & 16 & 32 & 64 & 128 & 256 \\
\midrule
 DNSP-MCMC & 5.56e-01 & 5.77e-01 & 5.78e-01 & 5.98e-01 & 6.02e-01 \\
\midrule
 DNSP-UNSP & 5.82e-01 & 5.92e-01 & 5.94e-01 & 5.98e-01 & 5.98e-01\\
 \midrule
 DNSP-USAP & 5.94e-01 & 5.94e-01 & 5.93e-01 & 5.93e-01 & 5.91e-01 \\
\bottomrule
\end{tabular}
\end{sc}
\end{small}
\end{table}

\begin{table}[H]
\caption{2-Hidden RMSE For Synthetic Dataset}
\label{table:2-hidden_rmse_synthetic}
\centering
\begin{small}
\begin{sc}
\begin{tabular}{cccccc}
\toprule
 \multirow{2}{*}{Model} & \multicolumn{5}{c}{Time(s)} \\ \cmidrule{2-6} & 500 & 900 & 2000 & 4000 & 8000 \\
\midrule
 DNSP-MCMC & 3.91e+08 & 2.38e+08 & 7.33e-01 & 6.95e-01 & 6.64e-01 \\
\midrule
 DNSP-UNSP & \textbackslash & 1.02e+01 & 9.44e+00 & 9.04e+00 & 8.97e+00\\
 \midrule
 DNSP-USAP & 3.44e+00 & 3.43e+00 & 3.30e+00 & 3.28e+00 & 3.28e+00 \\
\bottomrule
\end{tabular}
\end{sc}
\end{small}
\end{table}

\begin{table}[H]
\caption{2-Hidden Accuracy For Synthetic Dataset}
\label{table:2-hidden_accuracy_synthetic}
\centering
\begin{small}
\begin{sc}
\begin{tabular}{cccccc}
\toprule
 \multirow{2}{*}{Model} & \multicolumn{5}{c}{Time(s)} \\ \cmidrule{2-6} & 500 & 900 & 2000 & 4000 & 8000 \\
\midrule
 DNSP-MCMC & 8.08e-01 & 8.26e-01 & 8.48e-01 & 8.55e-01 & 8.68e-01 \\
\midrule
 DNSP-UNSP & \textbackslash & 8.30e-01 & 8.32e-01 & 8.36e-01 & 8.36e-01\\
 \midrule
 DNSP-USAP & 8.40e-01 & 8.40e-01 & 8.52e-01 & 8.60e-01 & 8.59e-01 \\
\bottomrule
\end{tabular}
\end{sc}
\end{small}
\end{table}

\begin{table}[H]
\caption{Results Of Baselines For Retweet Dataset}
\label{table:retweet_baseline}
\centering
\begin{small}
\begin{sc}
\begin{tabular}{ccc}
\toprule
Model & RMSE & Accuracy\\
\midrule
 THP & 1.56e+04 & 6.1e-01\\
 \midrule
 SAHP & 1.67e+04 & 4.5e-01\\
 \midrule
 NHP & 1.66e+04 & 4.9e-01\\
 \midrule
 MTPP & 1.66+e04 & 3.6e-01\\
\bottomrule
\end{tabular}
\end{sc}
\end{small}
\end{table}

\begin{table}[H]
\caption{1-Hidden RMSE For Retweet Dataset}
\label{table:1-hidden_rmse_retweet}
\centering
\begin{small}
\begin{sc}
\begin{tabular}{cccccc}
\toprule
 \multirow{2}{*}{Model} & \multicolumn{5}{c}{Time(s)} \\ \cmidrule{2-6} & 21500 & 43000 & 86000 & 172000 & 250000 \\
\midrule
 DNSP-MCMC & 1.45e+12 & 9.07e+11 & 4.52e+11 & 1.22e+11 & 5.05e+03 \\
\midrule
 DNSP-UNSP & 5.80e+10 & 9.65e+09 & 9.42e+03 & 9.40e+03 & 9.39e+03\\
 \midrule
 DNSP-USAP & 6.72e+10 & 1.76e+10 & 7.05e+03 & 7.05e+03 & 7.04e+03 \\
\bottomrule
\end{tabular}
\end{sc}
\end{small}
\end{table}

\begin{table}[H]
\caption{1-Hidden Accuracy For Retweet Dataset}
\label{table:1-hidden_accuracy_retweet}
\centering
\begin{small}
\begin{sc}
\begin{tabular}{cccccc}
\toprule
 \multirow{2}{*}{Model} & \multicolumn{5}{c}{Time(s)} \\ \cmidrule{2-6} & 21500 & 43000 & 86000 & 172000 & 250000 \\
\midrule
 DNSP-MCMC & 5.01e-01 & 5.33e-01 & 5.55e-01 & 5.71e-01 & 5.79e-01 \\
\midrule
 DNSP-UNSP & 5.11e-01 & 5.14e-01 & 5.15e-01 & 5.17e-01 & 5.17e-01\\
 \midrule
 DNSP-USAP & 5.30e-01 & 5.36e-01 & 5.41e-01 & 5.43e-01 & 5.43e-01 \\
\bottomrule
\end{tabular}
\end{sc}
\end{small}
\end{table}

\begin{table}[H]
\caption{2-Hidden RMSE For Retweet Dataset}
\label{table:2-hidden_rmse_retweet}
\centering
\begin{small}
\begin{sc}
\begin{tabular}{cccccc}
\toprule
 \multirow{2}{*}{Model} & \multicolumn{5}{c}{Time(s)} \\ \cmidrule{2-6} & 10300 & 20600 & 41200 & 82400 & 200000 \\
\midrule
 DNSP-MCMC & 9.93e+10 & 7.08e+10 & 4.84e+10 & 2.17e+10 & 3.98e+09 \\
\midrule
 DNSP-UNSP & 6.58e+10 & 1.24e+10 & 1.44e+09 & 9.42e+03 & \textbackslash\\
 \midrule
 DNSP-USAP & 2.66e+10 & 1.29e+10 & 4.76e+09 & 2.49e+09 & 6.28e+03 \\
\bottomrule
\end{tabular}
\end{sc}
\end{small}
\end{table}

\begin{table}[H]
\caption{2-Hidden Accuracy For Retweet Dataset}
\label{table:2-hidden_accuracy_retweet}
\centering
\begin{small}
\begin{sc}
\begin{tabular}{cccccc}
\toprule
 \multirow{2}{*}{Model} & \multicolumn{5}{c}{Time(s)} \\ \cmidrule{2-6} & 10300 & 20600 & 41200 & 82400 & 200000 \\
\midrule
 DNSP-MCMC & 6.36e-01 & 6.90e-01 & 7.30e-01 & 7.69e-01 & 7.87e-01 \\
\midrule
 DNSP-UNSP & 7.16e-01 & 7.20e-01 & 7.20e-01 & 7.21e-01 & \textbackslash\\
 \midrule
 DNSP-USAP & 6.89e-01 & 6.92e-01 & 6.95e-01 & 6.95e-01 & 6.97e-01 \\
\bottomrule
\end{tabular}
\end{sc}
\end{small}
\end{table}

\begin{table}[H]
\caption{Results Of Baselines For Earthquake Dataset}
\label{table:earthquake_baseline}
\centering
\begin{small}
\begin{sc}
\begin{tabular}{ccc}
\toprule
Model & RMSE & Accuracy\\
\midrule
 THP & 1.93e+03 & 6.0e-01\\
 \midrule
 SAHP & 1.79e+03 & 5.5e-01\\
 \midrule
 NHP & 1.95e+03 & 4.0e-01\\
 \midrule
 MTPP & 1.93e+03 & 5.1e-01\\
\bottomrule
\end{tabular}
\end{sc}
\end{small}
\end{table}

\begin{table}[H]
\caption{1-Hidden RMSE For Earthquake Dataset}
\label{table:1-hidden_rmse_earthquake}
\centering
\begin{small}
\begin{sc}
\begin{tabular}{cccccc}
\toprule
 \multirow{2}{*}{Model} & \multicolumn{5}{c}{Time(s)} \\ \cmidrule{2-6} & 2900 & 6200 & 11100 & 36200 & 60000 \\
\midrule
 DNSP-MCMC & 1.27e+12 & 7.56e+11 & 5.10e+09 & 1.26e+03 & 1.25e+03 \\
\midrule
 DNSP-UNSP & 2.03e+03 & 2.02e+03 & 2.02e+03 & 2.01e+03 & \textbackslash\\
 \midrule
 DNSP-USAP & 4.83e+09 & 1.74e+03 & 1.73e+03 & 1.73e+03 & 1.73e+03 \\
\bottomrule
\end{tabular}
\end{sc}
\end{small}
\end{table}

\begin{table}[H]
\caption{1-Hidden Accuracy For Earthquake Dataset}
\label{table:1-hidden_accuracy_earthquake}
\centering
\begin{small}
\begin{sc}
\begin{tabular}{cccccc}
\toprule
 \multirow{2}{*}{Model} & \multicolumn{5}{c}{Time(s)} \\ \cmidrule{2-6} & 2900 & 6200 & 11100 & 36200 & 60000 \\
\midrule
 DNSP-MCMC & 5.82e-01 & 5.90e-01 & 5.99e-01 & 6.02e-01 & 6.02e-01 \\
\midrule
 DNSP-UNSP & 6.02e-01 & 6.02e-01 & 6.02e-01 & 6.02e-01 & \textbackslash\\
 \midrule
 DNSP-USAP & 6.00e-01 & 6.01e-01 & 6.01e-01 & 6.02e-01 & 6.02e-01 \\
\bottomrule
\end{tabular}
\end{sc}
\end{small}
\end{table}

\begin{table}[H]
\caption{2-Hidden RMSE For Earthquake Dataset}
\label{table:2-hidden_RMSE_earthquake}
\centering
\begin{small}
\begin{sc}
\begin{tabular}{cccccc}
\toprule
 \multirow{2}{*}{Model} & \multicolumn{5}{c}{Time(s)} \\ \cmidrule{2-6} & 7200 & 14400 & 28800 & 57600 & 116016 \\
\midrule
 DNSP-MCMC & 4.92e+10 & 3.83e+10 & 2.55e+10 & 1.78e+10 & 1.44e+10 \\
\midrule
 DNSP-UNSP & 2.09e+03 & 2.09e+03 & 2.08e+03 & 2.08e+03 & \textbackslash\\
 \midrule
 DNSP-USAP & 2.16e+10 & 1.54e+10 & 1.53e+10 & 1.21e+09 & 1.42e+03 \\
\bottomrule
\end{tabular}
\end{sc}
\end{small}
\end{table}

\begin{table}[H]
\caption{2-Hidden Accuracy For Earthquake Dataset}
\label{table:2-hidden_accuracy_earthquake}
\centering
\begin{small}
\begin{sc}
\begin{tabular}{cccccc}
\toprule
 \multirow{2}{*}{Model} & \multicolumn{5}{c}{Time(s)} \\ \cmidrule{2-6} & 7200 & 14400 & 28800 & 57600 & 116016 \\
\midrule
 DNSP-MCMC & 6.05e-01 & 6.14e-01 & 6.22e-01 & 6.30e-01 & 6.48e-01 \\
\midrule
 DNSP-UNSP & 6.87e-01 & 6.87e-01 & 6.87e-01 & 6.87e-01 & \textbackslash\\
 \midrule
 DNSP-USAP & 6.04e-01 & 6.06e-01 & 6.06e-01 & 6.07e-01 & 6.07e-01 \\
\bottomrule
\end{tabular}
\end{sc}
\end{small}
\end{table}

\begin{table}[H]
\caption{Results Of Baselines For Homicide Dataset}
\label{table:homicide_baseline}
\centering
\begin{small}
\begin{sc}
\begin{tabular}{ccc}
\toprule
Model & RMSE & Accuracy\\
\midrule
 THP & 1.80e+05 & 2.96e-01\\
 \midrule
 SAHP & 1.80e+05 & 2.10e-01\\
 \midrule
 NHP & 1.80e+05 & 1.57e-01\\
 \midrule
 MTPP & 1.80e+05 & 2.05e-01\\
\bottomrule
\end{tabular}
\end{sc}
\end{small}
\end{table}

\begin{table}[H]
\caption{1-Hidden RMSE For Homicide Dataset}
\label{table:1-hidden_rmse_homicide}
\centering
\begin{small}
\begin{sc}
\begin{tabular}{cccccc}
\toprule
 \multirow{2}{*}{Model} & \multicolumn{5}{c}{Time(s)} \\ \cmidrule{2-6} & 483 & 966 & 1932 & 3864 & 5800 \\
\midrule
 DNSP-MCMC & 1.72e+05 & 1.47e+05 & 1.31e+05 & 1.21e+05 & 1.17e+05 \\
\midrule
 DNSP-UNSP & 1.56e+05 & 1.56e+05 & 1.55e+05 & 1.55e+05 & 1.55e+05\\
 \midrule
 DNSP-USAP & 1.43e+05 & 1.40e+05 & 1.40e+05 & 1.39e+05 & 1.40e+05 \\
\bottomrule
\end{tabular}
\end{sc}
\end{small}
\end{table}

\begin{table}[H]
\caption{1-Hidden Accuracy For Homicide Dataset}
\label{table:1-hidden_accuracy_homicide}
\centering
\begin{small}
\begin{sc}
\begin{tabular}{cccccc}
\toprule
 \multirow{2}{*}{Model} & \multicolumn{5}{c}{Time(s)} \\ \cmidrule{2-6} & 483 & 966 & 1932 & 3864 & 5800 \\
\midrule
 DNSP-MCMC & 2.02e-01 & 2.11e-01 & 2.11e-01 & 2.40e-01 & 2.55e-01 \\
\midrule
 DNSP-UNSP & 2.25e-01 & 2.42e-01 & 2.40e-01 & 2.49e-01 & 2.64e-01\\
 \midrule
 DNSP-USAP & 2.21e-01 & 2.44e-01 & 2.45e-01 & 2.56e-01 & 2.57e-01 \\
\bottomrule
\end{tabular}
\end{sc}
\end{small}
\end{table}

\begin{table}[H]
\caption{2-Hidden RMSE For Homicide Dataset}
\label{table:2-hidden_rmse_homicide}
\centering
\begin{small}
\begin{sc}
\begin{tabular}{cccc}
\toprule
 \multirow{2}{*}{Model} & \multicolumn{3}{c}{Time(s)} \\ \cmidrule{2-4} & 8300 & 16000 & 27000  \\
\midrule
 DNSP-MCMC & 1.77e+05 & 1.50e+05 & 1.37e+05 \\
\midrule
 DNSP-UNSP & 1.61e+05 & 1.60e+05 & 1.60e+05\\
 \midrule
 DNSP-USAP & 1.58e+05 & 1.58e+05 & 1.58e+05 \\
\bottomrule
\end{tabular}
\end{sc}
\end{small}
\end{table}

\begin{table}[H]
\caption{2-Hidden Accuracy For Homicide Dataset}
\label{table:2-hidden_accuracy_homicide}
\centering
\begin{small}
\begin{sc}
\begin{tabular}{cccc}
\toprule
 \multirow{2}{*}{Model} & \multicolumn{3}{c}{Time(s)} \\ \cmidrule{2-4} & 8300 & 16000 & 27000  \\
\midrule
 DNSP-MCMC & 3.32e-01 & 3.92e-01 & 4.44e-01 \\
\midrule
 DNSP-UNSP & 6.64e-01 & 6.72e-01 & 6.93e-01\\
 \midrule
 DNSP-USAP & 7.65e-01 & 8.17e-01 & 8.50e-01 \\
\bottomrule
\end{tabular}
\end{sc}
\end{small}
\end{table}
\subsection{Hardware and Software}
We run the experiments for synthetic datasets, retweets, and earthquakes in
a cluster. For each job, we use two cores from
a Intel\textregistered\
Xeon\textregistered\ Silver 4214
CPUs running at 2.20GHz and 1 GeForce\textregistered\  RTX 2080 Ti. 

We run each experiment for homicides in a machine with a core from
Intel\textregistered\ i7-5930K CPU and 1 GeForce\textregistered\ GTX 1080 Ti.

We use Pytorch to implement our algorithms.
\subsection{Implementation of MCMC for GPUs}
We re-implemented the MCMC algorithm to use GPUs. We do not randomly select a move from resampling, flip, and swap, instead, we have a deterministic order for these moves. We choose this setting to minimize the number of dimension changes of the tensors used to store the events. 

We store the real events and the virtual events in the same tensor. The only move that changes the dimension of this tensor is to re-sample the virtual events. For each MCMC sampling step, we first do a re-sampling, then followed by 3 flips, 1 swap, 3 flips, and 1 swap.
\section{SOCIETAL IMPACT}
Our proposed inference algorithm can be used to predict future earthquakes, violent crimes, and severe medical complications when speed is the top concern. The successful application of our method can help save a lot of lives. 

The misuse of our algorithm may also cause some problems. For example, some companies may use our algorithm to predict the customers' behavior and manipulate their buying activity, so customers may tend to buy more than what they need and then can cause some waste.

\end{document}